%% file: main.tex
\documentclass[10pt,twocolumn,letterpaper]{article}

\usepackage[pagenumbers]{cvpr} 

\input{preamble}

\definecolor{cvprblue}{rgb}{0.21,0.49,0.74}
\usepackage[pagebackref,breaklinks,colorlinks,allcolors=cvprblue]{hyperref}

\def\paperID{*****} 
\def\confName{CVPR}
\def\confYear{2026}

\title{ResTok: Learning Hierarchical Residuals in 1D Visual Tokenizers for Autoregressive Image Generation}

\author{Xu Zhang$^{1,2,}$\thanks{Work done while interning at Kuaishou Technology.} \quad \quad
Cheng Da$^{2}$ \quad \quad
Huan Yang$^{2,}$\thanks{Project leader.} \quad \quad
Kun Gai$^{2}$ \quad \quad
Ming Lu$^{1,}$\thanks{Corresponding author: {\tt \textless{}minglu@nju.edu.cn\textgreater{}}.} \quad \quad
Zhan Ma$^{1}$ \\
$^1$Vision Lab, Nanjing University \quad \quad $^2$Kolors Team, Kuaishou Technology
}

\begin{document}

\maketitle

\begin{abstract}
Existing 1D visual tokenizers for autoregressive (AR) generation largely follow the design principles of language modeling, as they are built directly upon transformers whose priors originate in language, yielding single-hierarchy latent tokens and treating visual data as flat sequential token streams. However, this language-like formulation overlooks key properties of vision, particularly the hierarchical and residual network designs that have long been essential for convergence and efficiency in visual models. To bring ``vision'' back to vision, we propose the \textbf{Res}idual \textbf{Tok}enizer (\textbf{ResTok}), a 1D visual tokenizer that builds hierarchical residuals for both image tokens and latent tokens. The hierarchical representations obtained through progressively merging enable cross-level feature fusion at each layer, substantially enhancing representational capacity. Meanwhile, the semantic residuals between hierarchies prevent information overlap, yielding more concentrated latent distributions that are easier for AR modeling. Cross-level bindings consequently emerge without any explicit constraints. To accelerate the generation process, we further introduce a hierarchical AR generator that substantially reduces sampling steps by predicting an entire level of latent tokens at once rather than generating them strictly token-by-token. Extensive experiments demonstrate that restoring hierarchical residual priors in visual tokenization significantly improves AR image generation, achieving a gFID of 2.34 on ImageNet-256 with only 9 sampling steps. Code is available at \url{https://github.com/Kwai-Kolors/ResTok}.
\end{abstract}
\vspace{-14pt}

\section{Introduction}
\label{sec:intro}

\begin{figure}[t]
\centering
\subfloat[Query along depth.]{\includegraphics[scale=0.41]{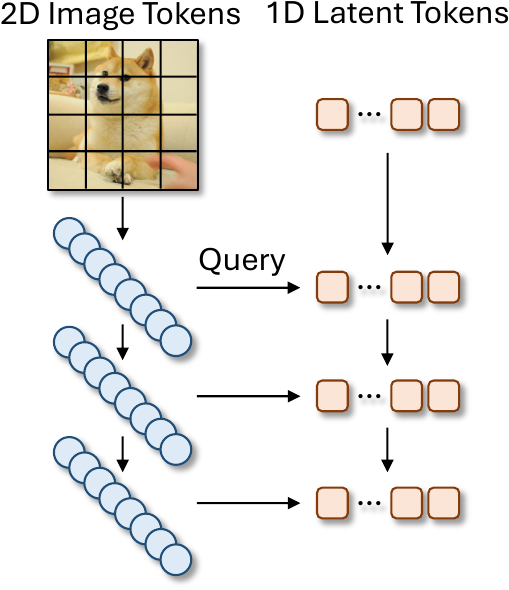}\label{subfig:titok}}
\subfloat[Query along depth and hierarchy.]{\includegraphics[scale=0.41]{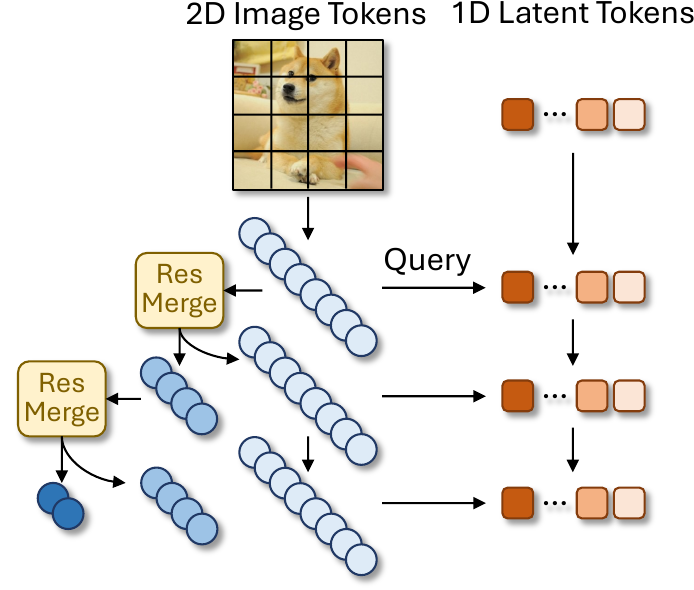}\label{subfig:restok}}
\caption{Comparison between (a) existing 1D tokenizers~\citep{yu2024titok,li2025imagefolder,huang2025spectralar,liu2025detailflow} querying features along only depth and (b) ResTok querying along both depth and hierarchy. By progressively merging image tokens, ResTok brings multi-scale hierarchies back to the ViT-based tokenizer, which encourages implicit alignments between image tokens and latent tokens and enforces better causalities of latent tokens for AR generation.}
\label{fig:teaser}
\vspace{-10pt}
\end{figure}

Autoregressive (AR) modeling has recently become a strong paradigm for high-quality visual generation and shows promise for unified multi-modal modeling. By predicting visual tokens sequentially, AR models inherit the scalability and controllability of language modeling. Their effectiveness, however, depends critically on how visual signals are tokenized, since tokenizers define the semantic dependencies AR models can learn and the reconstruction quality decoders can achieve. Auto-Encoding (AE)~\citep{hinton2006ae} naturally supports this process by learning compact latent representations. Its extensions, such as VAEs~\citep{kingma2014vae}, hierarchical VAEs~\citep{gregor2015draw,kingma2016iaf,sonderby2016ladder}, and VQ-VAEs~\citep{oord2017vqvae}, have substantially expanded representational capacity and become core components of modern generative models. Although pixel-level AR models~\citep{oord2016pixelrnn,oord2016pixelcnn,chen2020igpt} demonstrated strong performance, AE-based tokenizers remain essential for reducing dimensionality and capturing semantic structure. Contemporary frameworks therefore integrate AEs to improve fidelity and efficiency~\citep{esser2021vqgan,rombach2022ldm}. Within the Vision Transformer (ViT) paradigm~\citep{vaswani2017transformer,dosovitskiy2021vit}, this approach becomes particularly appealing, as images can be represented as sequences of latent tokens aligned with language-model-style training. As a result, tokenizer design emerges as a central challenge for further advancing AR visual generation.

To obtain 1D sequences for AR modeling, early visual tokenizers~\citep{esser2021vqgan,yu2022vitvqgan,lee2022rqvae} typically flattened 2D AE latents using raster scans or similar heuristics. Such strategies, however, are misaligned with AR causality at scan turning points where spatial continuity breaks down. To overcome this, later approaches abandon rigid spatial ordering and seek non-spatial token dependencies instead which are more compatible with AR modeling. Beyond multi-scale 2D tokenization~\citep{tian2024var}, another promising direction is 1D tokenization~\citep{ge2023seed,yu2024titok}. By discarding fixed spatial grids, query-based 1D tokenizers learn abstract semantics in a sequential form that aligns with AR prediction and resembles language modeling. Subsequent studies attempt to impose token causality by assigning levels to frequency bands~\citep{huang2025spectralar} or spatial resolutions~\citep{liu2025detailflow}, but such designs rely on non-semantic hand-crafted rules. Other methods introduce diffusion decoders to strengthen semantic learning~\citep{wen2025semanticist,bachmann2025flextok}, yet the dual stochastic processes (\ie, AR and diffusion) complicate optimization and lead to instability when scaling to longer token sequences.

Despite these advances, existing 1D tokenizers still face two main challenges:
(1) \textit{Lack of cross-level fusion}. Most methods~\citep{ge2023seed,yu2024titok,bachmann2025flextok,huang2025spectralar,xiong2025gigatok,liu2025detailflow} extract features from low- to high-level solely along network depth, but cannot fuse features from multiple levels at a certain layer. This is in contrast to feature-fusion studies~\citep{lin2017fpn,sun2019hrnet}, where cross-level fusion is known to be crucial for strong visual representation.
(2) \textit{High codebook entropy}. Since redundancy between latent tokens is rarely addressed, current approaches often produce similar embeddings in the codebook, yielding relatively uniform probabilities. Such high-entropy codebooks are unfriendly for AR modeling and may hinder generation performance.
We argue that these challenges stem from the ignorance of the intrinsic difference between vision and language. Existing methods adopt the same isotropic design as transformers, while vision properties like hierarchical residuals are gradually discarded as illustrated in \cref{fig:teaser}. To better uncover what enables efficient tokenization and generation, we introduce the \textbf{Res}idual \textbf{Tok}enizer (\textbf{ResTok}) and identify three key designs:
\begin{itemize}

\item \textbf{Hierarchical representations} enhance representational capacities, especially with multiple scales. To make the hierarchical design compatible with ViT-based tokenizers, we progressively merge image tokens into coarser features and insert them at the beginning of the token sequence. This allows latent tokens to fuse in-context features with image tokens across hierarchies.

\item \textbf{Semantic residuals} between hierarchies concentrate latent distributions. Unlike hand-crafted constraints~\citep{huang2025spectralar,liu2025detailflow} or additive residuals~\citep{tian2024var,li2025imagefolder}, ResTok learns residuals in a semantically structured way. By guiding the model to accumulate compensatory visual features, ResTok reduces the information overlap, resulting in lower-entropy codebooks that are easier for AR modeling.

\item \textbf{Accelerated generation} is enabled by proposing a hierarchical AR (HAR) variant of LlamaGen~\citep{sun2024llamagen} upon ResTok. Switching from next-token prediction to next-hierarchy prediction, the HAR generator significantly reduces sampling steps with acceptable degradation of generation performance.
\end{itemize}
By learning these visual properties, cross-level bindings emerge without explicit constraints: coarser latent tokens align with high-level image tokens, while finer latents capture low-level residual details. Coupled with LlamaGen-L~\citep{sun2024llamagen}, ResTok achieves state-of-the-art AR generation performance on the ImageNet 256$\times$256 benchmark~\citep{imagenet1k}, reaching a gFID of 2.34 with only 9 sampling steps.

\section{Related Work}

\subsection{Visual Tokenization}

Autoregressive visual generation hinges on effective tokenization. Early methods simply convert grid-based 2D latents from autoencoders into 1D sequences using raster scans~\citep{oord2017vqvae,esser2021vqgan,yu2022vitvqgan,lee2022rqvae,yu2024magvit2}. Innovations like SPAE~\citep{yu2023spae} explicitly aligns token hierarchies with semantic structures, underscoring the importance of cross-modal alignment. However, these approaches may disrupt autoregressive causality at scan turning points. To address this fundamental mismatch, query-based 1D visual tokenization techniques have emerged, which can learn naturally sequential tokens.

Notably, SEED~\citep{ge2023seed} and TiTok~\citep{yu2024titok} learn 1D latent sequences directly from image patches, aligning token order with abstract semantics rather than spatially matched tokens~\citep{beyer2025highly}. SpectralAR~\citep{huang2025spectralar} and DetailFlow~\citep{liu2025detailflow} further refine token causality by explicitly linking token length to frequency bands or spatial resolutions, encouraging shorter sequences to represent coarse visual features and longer ones to capture details. However, these methods rely on hand-crafted constraints, reducing flexibility. ImageFolder~\citep{li2025imagefolder} utilizes residual quantization~\citep{lee2022rqvae,tian2024var} with random drop of latent tokens to form a multi-scale latent scheme, but the hard additive residual design may not be optimal from the semantic perspective. In contrast, GigaTok~\citep{xiong2025gigatok} introduces latent hierarchies by applying progressive latent initialization at the input stage, while VFMTok~\citep{zheng2025vfmtok} directly uses learnable tokens to query single-scale visual features from multiple levels of a pre-trained foundation model.

\begin{figure*}[t]
\centering
\includegraphics[width=0.935\linewidth]{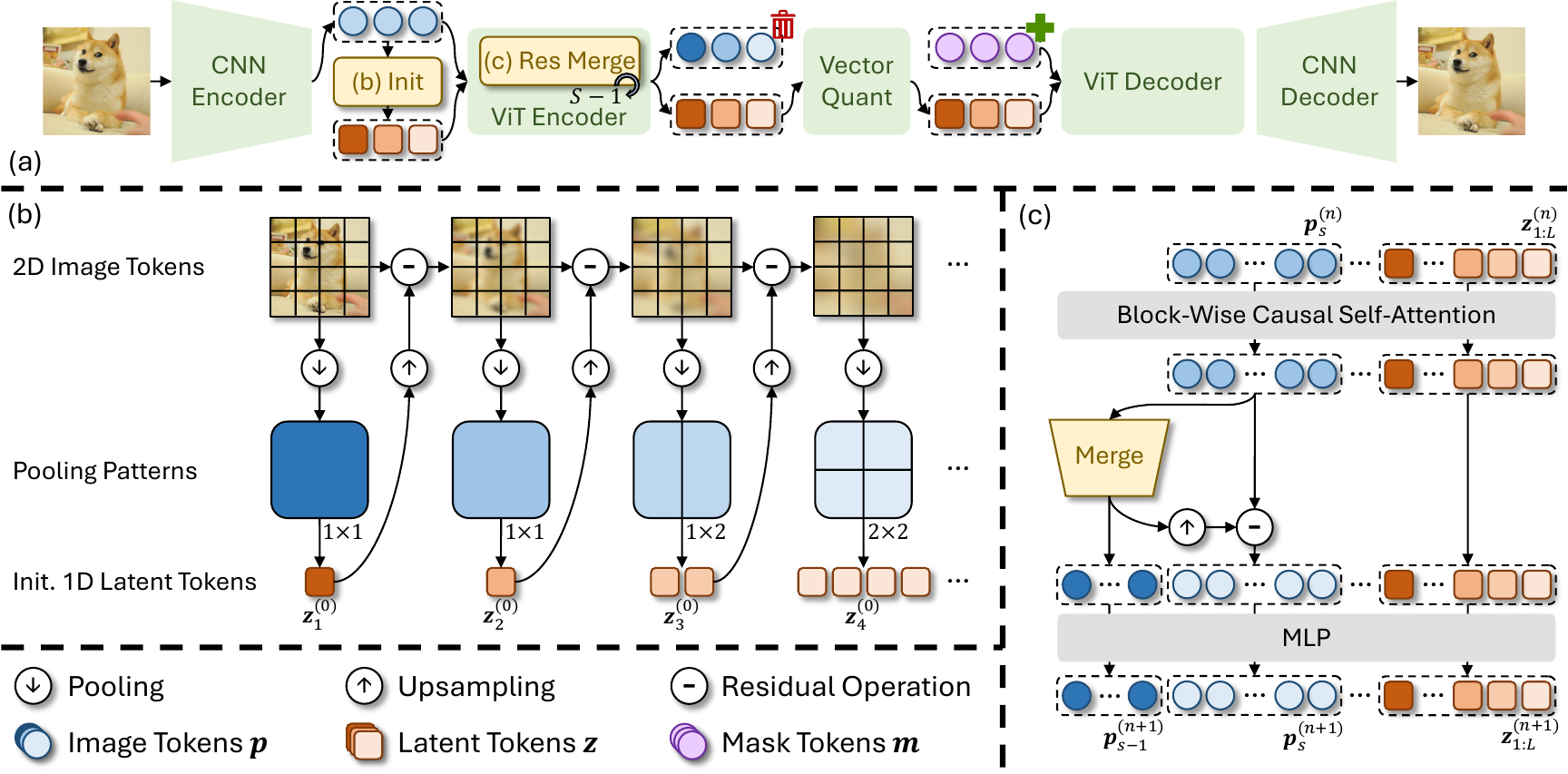}
\vspace{-6pt}
\caption{Overview of ResTok. (a) Pipeline of encoding and decoding processes. There are $S-1$ residual merging blocks uniformly replacing the original transformer blocks in the encoder, where $S$ denotes the number of scales. (b) Residual 1D latent token initialization. When increasing the target size of pooling, we first double the width, and then alternately double the height and width in subsequent steps. (c) Residual merging block. Average pooling is used as the merging method in our experiments.}
\label{fig:pipeline}
\vspace{-8pt}
\end{figure*}

\subsection{Autoregressive Image Generation}

In the realm of AR visual generation, foundational works begin with pixel-level AR models~\citep{oord2016pixelrnn,oord2016pixelcnn,chen2020igpt}, but these often struggle with efficiency due to high-dimensional input. More recent studies have shifted focus toward discrete latent token generation using VQ-VAE~\citep{oord2017vqvae} and its variants~\citep{esser2021vqgan,lee2022rqvae,tian2024var}, enabling powerful transformer-based AR models. VAR~\citep{tian2024var} introduces coarse-to-fine generation, while FlowAR~\citep{ren2025flowar} integrates flow matching~\citep{lipman2023flow} to model inter-scale dependencies. Infinity~\citep{han2025infinity} explores long-range refinement strategies for high-resolution generation. MaskGIT~\citep{chang2022maskgit} enables random prediction order, and MAR~\citep{li2024mar} eliminates the need of VQ for AR generation.

Despite these advances, the representative AR generation paradigm LlamaGen~\citep{sun2024llamagen} still attracts the main focus of the community, becoming the foundation of many following works~\citep{wang2025par,xiong2025gigatok,liu2025detailflow,zheng2025vfmtok}, as its simplicity and capability of integration with unified multi-modal models. Thus, in our work, we use LlamaGen as our testbed and propose a hierarchical variant for acceleration.

\section{Residual Tokenizer}
\label{sec:restok}

\subsection{Pipeline Overview}

In contrast to conventional 2D tokenizers~\citep{oord2017vqvae,esser2021vqgan,yu2022vitvqgan} used for AR generation, 1D tokenizers learn sequential latent tokens that query visual features directly from grid-structured image tokens. As shown in \cref{fig:pipeline}a, for the encoding process, given an input image $\bm{x} \in \mathbb{R}^{H \times W \times 3}$, a CNN encoder first transforms $\bm{x}$ into initial image tokens $\bm{p}^{(0)} \in \mathbb{R}^{\frac{H}{f} \times \frac{W}{f} \times C}$, downsampled by a factor of $f$. Here, the superscript $(0)$ denotes the input features of the ViT encoder or decoder, while $(n)$ later refers to the output features at the $n$-th transformer layer. The image tokens are then flattened and fed into a ViT encoder $\mathcal{E}(\cdot)$ together with a set of latent tokens $\bm{z}^{(0)}_{1:L}$ initialized from $\bm{p}^{(0)}$, where the subscript $1{:}L$ indicates the indices of the hierarchies. These latent tokens iteratively query and refine visual features across layers. After $N$ layers, the encoder outputs the final image tokens $\bm{p}^{(N)}$ and latent tokens $\bm{z}^{(N)}$. The latent tokens are quantized via $\bm{\hat{z}}^{(0)}_{1:L} = \text{VectorQuant}(\bm{z}^{(N)}_{1:L}; \mathcal{C})$, where $\mathcal{C}$ is the codebook, and the quantized latents $\bm{\hat{z}}^{(0)}_{1:L}$ serve as the representation used for reconstruction and generation. For the decoding process, a set of masked image tokens $\bm{m}^{(0)}_\text{img} \in \mathbb{R}^{\frac{H}{f} \times \frac{W}{f} \times C}$ initiates the ``inverse'' querying procedure. A ViT decoder $\mathcal{D}(\cdot)$ retrieves features from $\bm{\hat{z}}^{(0)}_{1:L}$ and outputs the restored image tokens $\bm{m}^{(N)}_\text{img}$. The reconstructed image $\bm{\hat{x}}$ is produced by a CNN decoder from $\bm{m}^{(N)}_\text{img}$.

\subsection{Hierarchical Representations in ViT}

As shown in \cref{subfig:titok}, previous works~\citep{yu2024titok,li2025imagefolder,bachmann2025flextok,xiong2025gigatok,huang2025spectralar,liu2025detailflow,zheng2025vfmtok} adopt single-hierarchy image tokens for tokenizers, limiting latent tokens to capturing hierarchical features from other levels. To this end, we propose progressive merging in isotropic ViT to learn hierarchical representations.

Akin to classical pyramid architectures~\citep{he2016resnet,lin2017fpn,sun2019hrnet}, intermediate features are progressively merged into smaller scales at specific layers, structuring multiple stages throughout the tokenizer. Specifically, we replace normal ViT blocks with residual merging blocks every $N/S$ layers except for the last layer as shown in \cref{fig:pipeline}c, where $N$ denotes the number of transformer depth and $S$ stands for the stage count. The multi-scale representations are denoted as $\{ \bm{p}_1, \ldots, \bm{p}_S \}$ in a coarse-to-fine order. At \textit{n}-th layer, after the self-attention operation, the \textit{s}-th-scale feature $\bm{p}^{(n)}_s$ is merged into a coarser scale $\bm{p}^{(n)}_{s-1}$. Compared to querying features along the transformer depth illustrated in \cref{fig:teaser}, this design makes the representations in ResTok across all scales accessible, which is beneficial to the hierarchical latent tokens for querying multi-level features.

Inspired by TiTok~\citep{yu2024titok}, we adopt in-context learning paradigm rather than the Q-Former~\citep{li2023blip2} architecture in GigaTok~\citep{xiong2025gigatok} and VFMTok~\citep{zheng2025vfmtok}, since image tokens should evolve through tokenization to progressively extract multi-scale features. Additionally, we apply encoder attention masks to restrict the coarser scales from accessing the finer scales, enforcing causalities across hierarchies of both image and latent tokens. Note that the decoder has no hierarchical design or attention mask for simplicity. We use average pooling as the merging operation in our experiments.

\begin{figure}[t]
\centering
\includegraphics[scale=0.6]{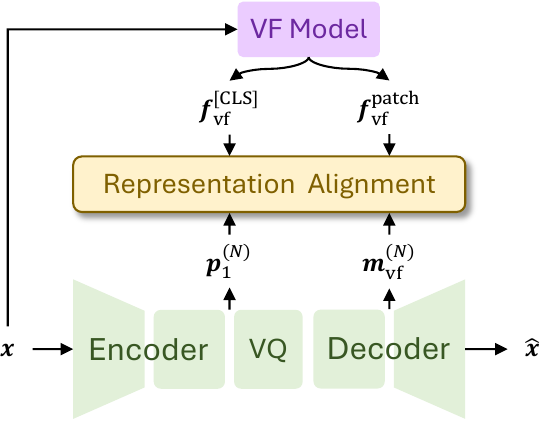}
\vspace{-4pt}
\caption{Representation alignment. The image $\bm{x}$ is processed by a VF model to get the \texttt{[CLS]} token $\bm{f}^\texttt{[CLS]}_\text{vf}$ and the visual tokens of image patches $\bm{f}^\text{patch}_\text{vf}$. The coarsest image tokens $\bm{p}^{(N)}_{1}$ and mask VF tokens $\bm{m}^{(N)}_\text{vf}$ are aligned with $\bm{f}^\texttt{[CLS]}_\text{vf}$ and $\bm{f}^\text{patch}_\text{vf}$, respectively.}
\label{fig:align}
\vspace{-10pt}
\end{figure}

\subsection{Semantic Residuals}

Some studies~\citep{xiong2025gigatok,zheng2025vfmtok} introduce multi-level image or latent tokens by naively stacking visual representations, but they often overlook the substantial information overlap between levels. This redundancy produces similar codebook embeddings and high entropy, which is unfavorable for AR modeling. Although methods such as VAR~\citep{tian2024var} and ImageFolder~\citep{li2025imagefolder} add residuals at the quantization bottleneck, these residuals are not accumulated semantically along the token sequence and thus fail to bind clear semantic attributes to latent tokens. To address these issues, we propose semantic residuals for both image and latent tokens.

For latent tokens, we apply residual initialization at the input stage. As shown in \cref{fig:pipeline}b, the number of latent tokens increases exponentially across hierarchical levels, except for the first two levels~\citep{xiong2025gigatok}. This results in a nested growth of token length across levels. To introduce residuals on top of hierarchical latent tokens, we do not always pool the feature map $\bm{p}^{(0)}$ directly to each target level length. Instead, inspired by the iterative approach in VAR~\citep{tian2024var}, we upsample the pooled feature back to the original size of $\bm{p}^{(0)}$, subtract $\bm{p}^{(0)}$ from the upsampled feature to obtain the residual, and then pool the residual to generate latent tokens. This residual formulation provides an initial guidance during training and prevents excessive information overlap among latent tokens. Similar operations are also been done for image tokens. At \textit{n}-th layer, $\bm{p}^{(n)}_s$ is subtracted from the upsampled $\bm{p}^{(n)}_{s-1}$ to obtain the residual relative to $\bm{p}^{(n)}_{s-1}$ rather than keeping the original image tokens in the sequence as shown in \cref{fig:pipeline}c.

\begin{figure}[t]
\centering
\includegraphics[width=1.0\linewidth]{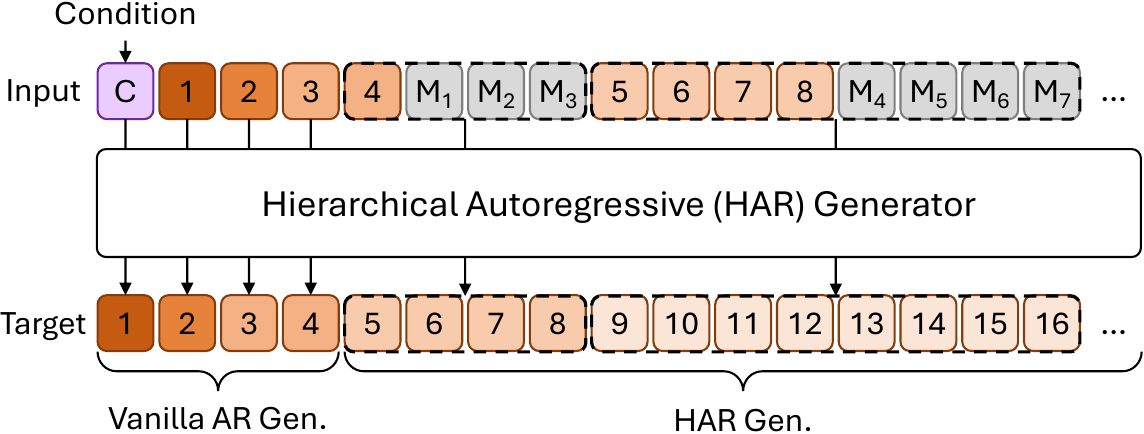}
\caption{Hierarchical autoregressive generator. The numbers in the colored tokens stand for the indices of the latent tokens. \texttt{[M\textsubscript{\textnormal{\textit{i}}}]} denotes the mask token filled at the \textit{i}-th missing position.}
\label{fig:har}
\vspace{-10pt}
\end{figure}

\begin{table*}[t]
\centering
\caption{System-level comparison of reconstruction and class-conditional generation on ImageNet 256$\times$256. ``Mask.'' and ``Diff.'' stand for masked generation and diffusion. ``\#Tokens'': the number of tokens needed to represent an image. ``\#Steps'': the number of sampling steps needed for generation. \dag: Training set includes data besides ImageNet. \ddag: Without classifier-free guidance. $\diamond$: Tokenizers are initialized with pre-trained vision foundation models. $\triangledown$: Images are downsampled from larger sizes than 256$\times$256. $\star$: Results are of 32 tokens.}
\label{tab:main_cmp}
\resizebox{1.0\linewidth}{!}{%
\begin{tabular}{llllllllllll}
\toprule
\multicolumn{1}{l|}{\multirow{2}{*}{Method}}                                  & \multicolumn{4}{l|}{Tokenizer}                                   & \multicolumn{7}{l}{Generator}                                             \\ \cmidrule(l){2-12}
\multicolumn{1}{l|}{}                                                         & Type & \#Param. & \#Tokens & \multicolumn{1}{l|}{rFID↓}          & Type        & \#Param. & \#Steps & gFID↓          & IS↑   & Pre.↑ & Rec.↑ \\ \midrule
\multicolumn{12}{c}{\textit{Continuous Token Modeling}}                                                                                                                                                                   \\ \midrule
\multicolumn{1}{l|}{LDM-4-G~\citep{rombach2022ldm}}                           & KL   & 55M      & 4096     & \multicolumn{1}{l|}{0.27$^\dag$}    & Diff.       & 400M     & 250     & 3.60           & 247.7 & -     & -     \\
\multicolumn{1}{l|}{DiT-XL/2~\citep{peebles2023dit}}                          & KL   & 84M      & 1024     & \multicolumn{1}{l|}{0.62$^\dag$}    & Diff.       & 675M     & 250     & 2.27           & 278.2 & 0.83  & 0.57  \\
\multicolumn{1}{l|}{LightningDiT-XL~\citep{yao2025vavae}}                     & KL   & 70M      & 256      & \multicolumn{1}{l|}{0.28}           & Diff.       & 675M     & 250     & 1.35           & 295.3 & 0.79  & 0.65  \\
\multicolumn{1}{l|}{MAR-B~\citep{li2024mar}}                                  & KL   & 66M      & 256      & \multicolumn{1}{l|}{0.87}           & Mask.+Diff. & 208M     & 64      & 2.31           & 281.7 & 0.82  & 0.57  \\
\multicolumn{1}{l|}{FlowAR-B~\citep{ren2025flowar}}                           & KL   & 66M      & 256      & \multicolumn{1}{l|}{0.87}           & VAR+Flow    & 300M     & 5       & 2.90           & 272.5 & 0.84  & 0.54  \\ \midrule
\multicolumn{12}{c}{\textit{Discrete Token Modeling}}                                                                                                                                                                     \\ \midrule
\multicolumn{12}{l}{\textit{Grid-Based Tokenization}}                                                                                                                                                                     \\
\multicolumn{1}{l|}{VQGAN~\citep{esser2021vqgan}}                             & VQ   & 23M      & 256      & \multicolumn{1}{l|}{4.98}           & AR          & 1.4B     & 256     & 15.78$^\ddag$  & 74.3  & -     & -     \\
\multicolumn{1}{l|}{RQTran.~\citep{lee2022rqvae}}                             & RQ   & 66M      & 256      & \multicolumn{1}{l|}{3.20}           & AR          & 3.8B     & 68      & 7.55$^\ddag$   & 134.0 & -     & -     \\
\multicolumn{1}{l|}{MaskGIT~\citep{chang2022maskgit}}                         & VQ   & 66M      & 256      & \multicolumn{1}{l|}{2.28}           & Mask.       & 227M     & 8       & 6.18$^\ddag$   & 182.1 & 0.80  & 0.51  \\
\multicolumn{1}{l|}{VAR-$d$16~\citep{tian2024var}}                            & MSRQ & 109M     & 680      & \multicolumn{1}{l|}{0.90$^\dag$}    & VAR         & 310M     & 10      & 3.30           & 274.4 & 0.84  & 0.51  \\
\multicolumn{1}{l|}{LlamaGen-L$^\triangledown$~\citep{sun2024llamagen}}       & VQ   & 72M      & 576      & \multicolumn{1}{l|}{0.94}           & AR          & 343M     & 576     & 3.07           & 256.1 & 0.83  & 0.52  \\
\multicolumn{1}{l|}{PAR-L-4$\times$$^\triangledown$~\citep{wang2025par}}      & VQ   & 72M      & 576      & \multicolumn{1}{l|}{0.94}           & PAR         & 343M     & 147     & 3.76           & 218.9 & 0.84  & 0.50  \\
\multicolumn{1}{l|}{IBQ-B~\citep{shi2025ibq}}                                 & IBQ  & 128M     & 256      & \multicolumn{1}{l|}{1.37}           & AR          & 342M     & 256     & 2.88           & 254.7 & 0.84  & 0.51  \\ \midrule
\multicolumn{12}{l}{\textit{Query-Based Tokenization}}                                                                                                                                                                    \\
\multicolumn{1}{l|}{TiTok-L-32~\citep{yu2024titok}}                           & VQ   & 641M     & 32       & \multicolumn{1}{l|}{2.21}           & Mask.       & 177M     & 8       & 2.77           & 199.8 & -     & -     \\
\multicolumn{1}{l|}{FlexTok d18-d18~\citep{bachmann2025flextok}}              & FSQ  & 950M     & 1-256    & \multicolumn{1}{l|}{1.61$^{\star}$} & AR+Flow     & 1.33B    & 26-281  & 2.02$^{\star}$ & -     & -     & -     \\
\multicolumn{1}{l|}{ImageFolder$^\diamond$~\citep{li2025imagefolder}}         & MSRQ & 176M     & 286      & \multicolumn{1}{l|}{0.80}           & VAR         & 362M     & 10      & 2.60           & 295.0 & 0.75  & 0.63  \\
\multicolumn{1}{l|}{GigaTok-B-L~\citep{xiong2025gigatok}}                     & VQ   & 622M     & 256      & \multicolumn{1}{l|}{0.81}           & AR          & 111M     & 256     & 3.26           & 221.0 & 0.81  & 0.56  \\
\multicolumn{1}{l|}{SpectralAR-$d$16~\citep{huang2025spectralar}}             & VQ   & -        & 64       & \multicolumn{1}{l|}{4.03}           & AR          & 310M     & 64      & 3.02           & 282.2 & 0.81  & 0.55  \\
\multicolumn{1}{l|}{DetailFlow-16$^\diamond$~\citep{liu2025detailflow}}       & VQ   & 271M     & 128      & \multicolumn{1}{l|}{1.22}           & PAR         & 326M     & 23      & 2.96           & 221.4 & 0.82  & 0.57  \\
\multicolumn{1}{l|}{VFMTok$^{\diamond\triangledown}$~\citep{zheng2025vfmtok}} & VQ   & -        & 256      & \multicolumn{1}{l|}{0.89}           & AR          & 343M     & 256     & 2.75           & 278.8 & 0.84  & 0.57  \\
\rowcolor{lightgray}\multicolumn{1}{l|}{ResTok (Ours)}                        & VQ   & 662M     & 128      & \multicolumn{1}{l|}{1.28}           & HAR         & 326M     & 9       & 2.34           & 257.8 & 0.79  & 0.60  \\ \bottomrule 
\end{tabular}%
}
\vspace{-3pt}
\end{table*}

\subsection{Optimization Strategies}

Representation alignment~\citep{yu2025repa,yao2025vavae} with a pre-trained vision foundation (VF) model is incorporated in ResTok for faster convergence. Different from existing aligned 1D tokenizers~\citep{liu2025detailflow,zheng2025vfmtok}, we apply alignment to both the encoder and the decoder as shown in \cref{fig:align}. At the encoder side, we apply global average pooling to the coarsest output hierarchy of image tokens $\bm{p}^{(N)}_{1}$ and align it to the \texttt{[CLS]} token of DINOv3-L~\citep{simeoni2025dinov3} via a linear layer $\phi_\text{enc}(\cdot)$ and \cref{eq:loss_enc} to guide the residual merging process. At the decoder side, we double the training batch, replace half of the mask image tokens $\bm{m}^{(0)}_\text{img}$ with mask VF tokens $\bm{m}^{(0)}_\text{vf}$~\citep{zheng2025vfmtok}, and align the corresponding output $\bm{m}^{(N)}_\text{vf}$ with the visual tokens of DINOv3-L~\citep{simeoni2025dinov3} through a linear layer $\phi_\text{dec}(\cdot)$ and \cref{eq:loss_dec}, which can preserve semantics at the quantization bottleneck. The VF loss $\mathcal{L}_\text{vf}$ can be formally written as
\begin{align}
\mathcal{L}_\text{enc} &= \text{ReLU}(\delta_\text{enc} - \text{CosSim}(\bm{p}^{(N)}_1), \phi_\text{enc}(\bm{f}^\texttt{[CLS]}_\text{vf}))),\label{eq:loss_enc}\\
\mathcal{L}_\text{dec} &= \text{ReLU}(\delta_\text{dec} - \text{CosSim}(\bm{m}^{(N)}_\text{vf}, \phi_\text{dec}(\bm{f}^\text{patch}_\text{vf}))),\label{eq:loss_dec}\\
\mathcal{L}_\text{vf} &= \lambda_\text{enc}\mathcal{L}_\text{enc} + \lambda_\text{dec}\mathcal{L}_\text{dec},\label{eq:loss_vf}
\end{align}
where $\text{ReLU}(\cdot)$ and $\text{CosSim}(\cdot,\cdot)$ denote clamping and cosine similarity, respectively. $\lambda_\text{enc}$ and $\lambda_\text{dec}$ control the trade-off between $\mathcal{L}_\text{enc}$ and $\mathcal{L}_\text{dec}$. We set margins $\delta_\text{enc}$ and $\delta_\text{dec}$ in \cref{eq:loss_enc,eq:loss_dec} to control the similarities~\citep{yao2025vavae}, both fixed to 0.85 across experiments. Ablations in \cref{sec:ablation} validate the effectiveness of this co-design of $\mathcal{L}_\text{vf}$.

To keep ResTok simple, we do not tie the latent tokens to manually decided spatial resolutions~\citep{liu2025detailflow} or frequency bands~\citep{huang2025spectralar}. Instead, we optimize each latent hierarchy to the same training objectives \cref{eq:loss_total} with commonly used MSE loss $\mathcal{L}_\text{mse}$, perceptual loss~\citep{zhang2018lpips} $\mathcal{L}_\text{percp}$, GAN loss~\citep{ian2014gan} $\mathcal{L}_\text{gan}$ and VF loss $\mathcal{L}_\text{vf}$:
\begin{equation}
\mathcal{L}_\text{total} = \lambda_\text{mse}\mathcal{L}_\text{mse} + \lambda_\text{percp}\mathcal{L}_\text{percp} + \lambda_\text{gan}\mathcal{L}_\text{gan} + \lambda_\text{vf}\mathcal{L}_\text{vf},\label{eq:loss_total}
\end{equation}
where $\lambda_\text{mse}$, $\lambda_\text{percp}$, $\lambda_\text{gan}$ and $\lambda_\text{vf}$ balance the loss terms, making the tokenizer adaptively and implicitly decide the optimal visual features of a certain length. This implicit method can also encourage semantic accumulation along the residual token sequence rather than non-semantic information.

Moreover, we do not explicitly tie any latent token group to a certain image hierarchy, which encourages self-alignment of image and latent hierarchies. To further promote this self-alignment property, we apply nested dropout of latent hierarchies~\citep{miwa2025onedpiece,bachmann2025flextok,li2025imagefolder,liu2025detailflow}, which can guide the tokenizer to learn essential visual features needed for reconstruction at each semantic level, aligning with our multi-scale hierarchical designs.

\begin{figure*}[t]
\centering
\includegraphics[width=0.955\linewidth]{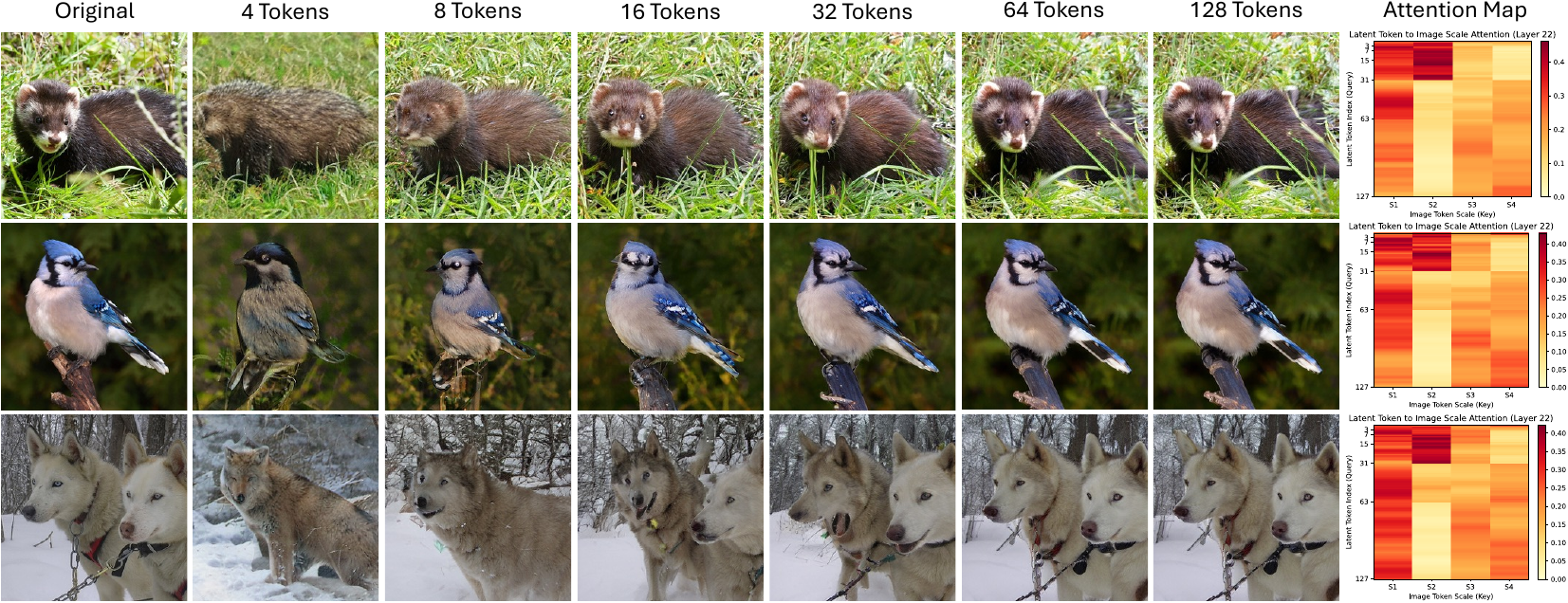}
\vspace{-4pt}
\caption{Visualizations of reconstructions with various token lengths and attention weights in the encoder. The first 16 latent tokens are more closely associated with the coarser image scales S1 and S2, capturing high-level semantics (\eg, object, position, color, etc.). In contrast, the subsequent latent tokens progressively refine fine-grained details, primarily querying the finer image tokens from S3 and S4.}
\label{fig:recon}
\vspace{-10pt}
\end{figure*}

\section{Hierarchical Autoregressive Generation}

The original LlamaGen~\citep{sun2024llamagen} adopts the next-token prediction (NTP) paradigm, hindering the generation speed with long sequences. While ResTok is capable of NTP, we also develop a hierarchical autoregressive (HAR) generator tailored to ResTok's hierarchical design to further boost the speed of AR generation.

As illustrated in \cref{fig:har}, the generation process can be divided into two parts, vanilla AR generation and HAR generation. In the vanilla AR generation phase, a group of latent tokens is predicted in an NTP manner. These tokens perform as initialization for the following HAR prediction, reducing accumulation of sampling error in the beginning~\citep{liu2025detailflow}. In the HAR generation phase, the first HAR group has only one predicted token accompanied with special mask tokens, whose sum equals to the number of tokens in the next hierarchy of ResTok. Different from PAR~\citep{wang2025par} and DetailFlow~\citep{liu2025detailflow}, each hierarchy in ResTok has a different number of latent tokens, so we need to add mask tokens to each group to reach the next hierarchy's token count. In the training process, a hierarchical grouped attention mask is applied, while the optimization objective remains the same as LlamaGen~\citep{sun2024llamagen}. In our experiments, the number of NTP tokens equals to the number of minimal remaining tokens in nested token dropout training~\citep{miwa2025onedpiece,bachmann2025flextok,li2025imagefolder,liu2025detailflow}.

\section{Experiments}

\subsection{Experimental Settings}
\label{sec:setting}

\textbf{Implementation Details.} ResTok builds on TiTok-L~\citep{yu2024titok}, incorporating 128 latent tokens, a codebook $\mathcal{C}$ with 8,192 entries and a dimension of 8, a CNN encoder-decoder pair~\citep{xiong2025gigatok}, nested token dropout~\citep{miwa2025onedpiece,bachmann2025flextok,li2025imagefolder,liu2025detailflow} (the number of minimal remaining tokens is set to 4), a DINO discriminator~\citep{tian2024var}, and M-RoPE~\citep{wang2024qwen2vl}. These updates yield a strong baseline for the proposed modules in \cref{sec:restok} and our ablation study. For the main results, ResTok is trained on ImageNet training set~\citep{imagenet1k} at 256$\times$256 for 200 epochs with adversarial training beginning at step 20K, and LlamaGen-L~\citep{sun2024llamagen} is trained under HAR scheme for 300 epochs. For the ablations, ResTok and LlamaGen-L are trained on ImageNet for 30 epochs and 50 epochs, respectively. For both tokenizer and generator, we use a batch size of 256, AdamW optimizer~\citep{loshchilov2019adamw}, an initial learning rate of $1 \times 10^{-4}$ with one-epoch linear warm-up, and cosine decay to $1 \times 10^{-5}$ thereafter. In our experiments, all merging, pooling and upsampling operations use nearest interpolation. More details can be found in \cref{appx:impl}.

\noindent\textbf{Evaluation Metrics.} We utilize Fréchet Inception Distance (FID)~\citep{heusel2017fid}, Inception Score (IS)~\citep{salimans2016is}, Precision, and Recall as metrics for assessing reconstruction and generation performance. Since all of the ResTok variants in the ablation study achieve 100\% codebook utilization, we report the codebook entropy $H_\mathcal{C}$ instead as a better indicator to examine how various settings affect the concentration of the latent distribution and its correlation with FID.

\begin{figure*}[t]
\centering
\includegraphics[width=0.95\linewidth]{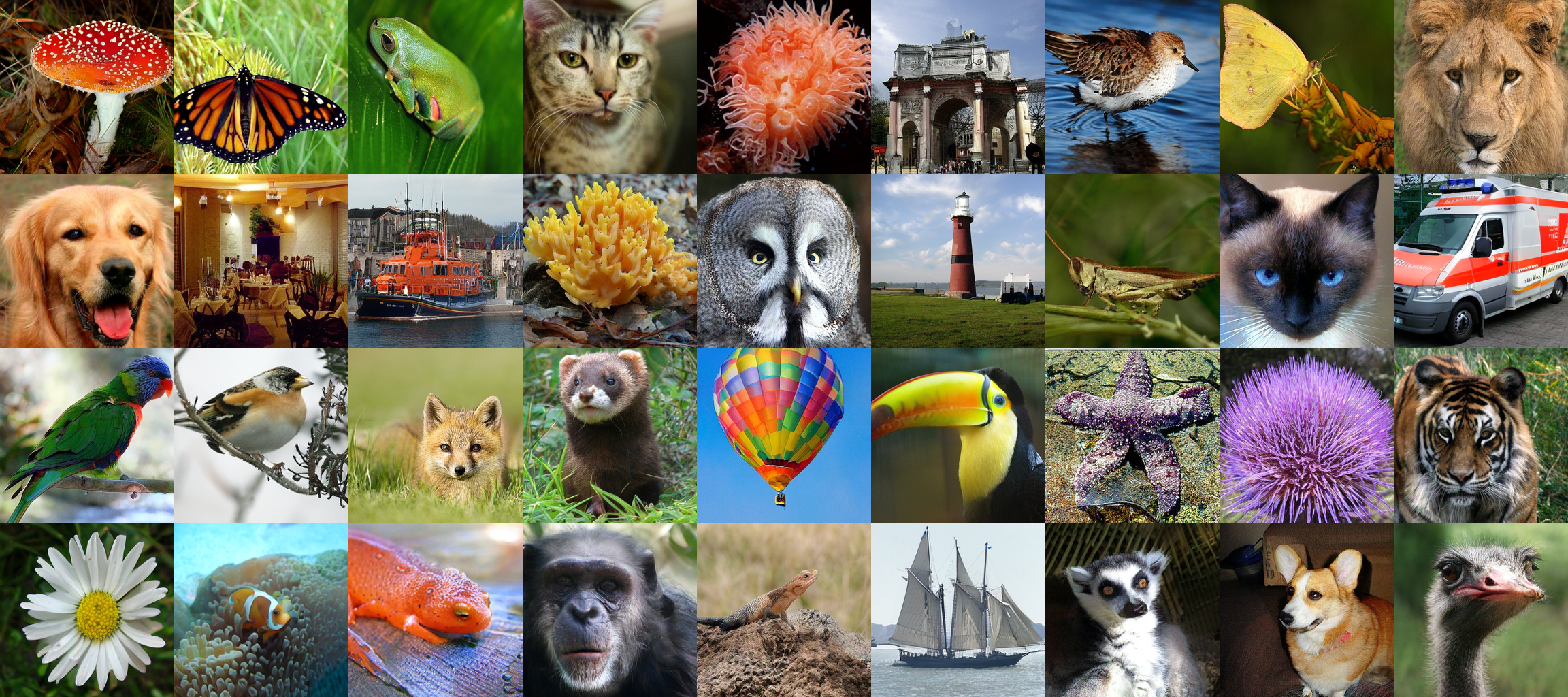}
\caption{Visualizations of generated 256$\times$256 samples on ImageNet-1K. By enhancing the representation capabilities of the tokenizer and constraining the causal dependencies among latent tokens, ResTok enables the AR generator to produce high-quality and diverse images.}
\label{fig:gen}
\vspace{-4pt}
\end{figure*}

\subsection{Quantitative Results}
\label{sec:quantitative_results}

We compare the proposed ResTok with recent representative methods across continuous and discrete token modeling paradigms in \cref{tab:main_cmp}. From the perspective of discrete methods, query-based visual tokenizers generally achieve better gFID, often reaching below 3.0 gFID with a $\sim$300M generator. Meanwhile, rFID remains competitive when scaling up model capacity and latent sequence length, with around 128 latent tokens typically enabling rFID scores near 1.0. This trend highlights that query-based tokenizers align more naturally with AR image generation.

Among query-based tokenizers, ResTok enables the accelerated HAR generator to achieve a state-of-the-art 2.34 gFID with only 9-step sampling, outperforming both prior query-based methods with stronger rFID~\citep{li2025imagefolder,xiong2025gigatok,zheng2025vfmtok} and other accelerated AR models that rely on longer latent sequences~\citep{tian2024var,wang2025par,li2025imagefolder,liu2025detailflow}. More concretely, although ResTok's rFID is slightly higher than DetailFlow~\citep{liu2025detailflow}, which also uses 128 latent tokens, ResTok benefits from its semantically organized codebook, enabling easier AR modeling and significantly improving gFID while requiring far fewer sampling steps. Compared to ImageFolder~\citep{li2025imagefolder}, ResTok attains better gFID and sampling efficiency, yet uses only 128 latent tokens instead of 286, demonstrating a substantially more compact and efficient representation. Furthermore, despite operating under a pure AR framework, ResTok and HAR remain competitive with recent hybrid (masked) AR and diffusion methods~\citep{li2024mar,ren2025flowar,bachmann2025flextok}, highlighting the effectiveness of reinstating hierarchical residual priors in 1D visual tokenization.

\subsection{Qualitative Results}

By learning semantic hierarchical residuals, ResTok exhibits a coherent semantic stacking behavior as shown in \cref{fig:recon}. The model reconstructs images in a coarse-to-fine manner where each additional group of latent tokens contributes semantically meaningful refinements, such as object identity, spatial layout, color composition, and finally textural and boundary details. This is distinctly different from SpectralAR~\citep{huang2025spectralar} and DetailFlow~\citep{liu2025detailflow}, where the refinement stages primarily operate on frequency bands or low-level textures without establishing clear semantic ordering. The emergent property observed in ResTok suggests that its latent tokens are more aligned with semantic attributes, enabling more controllable generation.

To further understand the underlying mechanisms of hierarchical residuals in ResTok, we visualize the encoder attention maps in \cref{fig:recon}. By comparing the reconstructed images from different token lengths with their corresponding attention maps, we can observe a clear alignment between the scales of image tokens and the represented content. The first 16 latent tokens primarily encode abstract semantic information, which corresponds to the coarser image scales $\bm{p}_1$ and $\bm{p}_2$ (\ie, S1 and S2 in \cref{fig:recon}). As the token sequence progresses, the later latent tokens gradually refine fine-grained details, mainly supported by the finer image scales $\bm{p}_3$ and $\bm{p}_4$ (\ie, S3 and S4 in \cref{fig:recon}). Additionally, the attention maps in \cref{fig:recon} show that the coarsest scale S1 of image tokens act as a global semantic source, which the latent tokens query most. The rest scales of image tokens compensate residuals to the latent tokens, naturally exhibiting a coarse-to-fine transition property. It reveals that the hierarchical residual properties are essential for the tokenizer to capture information at distinct semantic levels.

Such latent tokens organized by semantics with a low-entropy codebook are also more amenable to modeling by the AR generator, such as LlamaGen~\citep{sun2024llamagen}, enabling high-quality and diverse image generation as shown in \cref{fig:gen}.

\subsection{Ablation Study}
\label{sec:ablation}

To thoroughly analyze the effectiveness of the proposed modules in ResTok, we conduct a series of ablations based on the improved baseline as described in \cref{sec:setting}. Unless otherwise specified, gFID is generated by vanilla AR generation without classifier-free guidance (CFG)~\citep{ho2021cfg}.

\begin{table}[t]
\centering
\caption{Ablation study on the network designs. The pooling factors of hierarchical image tokens are fixed to 2 by default.}
\vspace{-4pt}
\label{tab:abl_hiera_res}
{\small
\begin{tabular}{@{}l|l|lll@{}}
\toprule
ID & Setting                                  & rFID↓            & gFID↓            & $H_\mathcal{C}$ \\ \midrule
1  & Baseline                                 & 1.87             & 6.01             & 11.89           \\
2  & \hspace{1ex}+ Hierarchical Latent Tokens & 1.86             & 5.39             & 11.90           \\ \midrule
   & \hspace{1ex}+ Hierarchical Image Tokens  &                  &                  &                 \\
3  & \hspace{3ex}  2 Hiera.                   & 1.71             & 5.41             & 12.12           \\
4  & \hspace{3ex}  3 Hiera.                   & 1.70             & 5.53             & 11.91           \\
5  & \hspace{3ex}  4 Hiera. (default)         & \underline{1.67} & 6.58             & 11.47           \\ \midrule \midrule
   & \hspace{1ex}+ Residual Tokens            &                  &                  &                 \\
6  & \hspace{3ex}  Image Tokens               & 1.86             & 5.64             & 11.58           \\ 
7  & \hspace{3ex}  Latent Tokens              & 2.02             & 4.78             & 10.58           \\
8  & \hspace{3ex}  Both (default)             & 2.11             & \underline{4.56} & 8.79            \\ \bottomrule
\end{tabular}%
}
\vspace{-6pt}
\end{table}

\noindent \textbf{Hierarchical Residuals.} We begin with the network designs of hierarchical residuals, resulting in \cref{tab:abl_hiera_res}. The principles can be roughly divided into two parts: hierarchies and residuals. The former enhances representation capabilities for better reconstruction, and the latter concentrates latent distributions for lower gFID. Applying hierarchies to latent tokens (\ie, setting \#2) explicitly enforces the causality, improving gFID over the baseline even without residuals. Further adding hierarchies to image tokens (\ie, settings \#3 to \#5) significantly boosts the performance of reconstruction. By ablating the number of hierarchies, we find that the tokenizer with 4 hierarchies, which is also a typical configuration of conventional hierarchical neural networks~\citep{he2016resnet,liu2021swin,liu2022convnext}, strikes a balance between rFID and complexity. Then we explore the most suitable residual settings, \ie, settings \#6 to \#8. It shows that applying residuals to image tokens and latent tokens simultaneously performs best, with the lowest codebook entropy $H_\mathcal{C}$ and gFID.

We also ablate the best pooling factor of residual merging in \cref{fig:pipeline}c. \cref{tab:abl_hiera_pooling} reveals that merging image tokens with a pooling factor of 2 yields the best generation performance among the tested settings. This configuration provides a moderate level of abstraction compared with no pooling, while avoiding the excessive semantic loss at the smallest scale of image tokens observed with a 4$\times$ pooling.

\begin{table}[t]
\centering
\caption{Ablation study on the pooling factor in all hierarchies of image tokens. The number of hierarchies is set to 4 by default.}
\vspace{-4pt}
\label{tab:abl_hiera_pooling}
{\small
\begin{tabular}{@{}l|lll@{}}
\toprule
Pooling Factor  & rFID↓ & gFID↓             & $H_\mathcal{C}$ \\ \midrule
1 (w/o Pooling) & 1.89  & 5.81              & 10.32           \\
2 (Default)     & 2.11  & \underline{4.56}  & 8.79            \\
4               & 1.90  & 4.70              & 10.17           \\ \bottomrule
\end{tabular}%
}
\end{table}

\begin{table}[t]
\centering
\caption{Ablation study on the alignment positions.}
\vspace{-4pt}
\label{tab:abl_align}
{\small
\begin{tabular}{@{}ll|lll@{}}
\toprule
\multicolumn{2}{@{}l|}{Alignment Position} & \multirow{2}{*}{rFID↓} &\multirow{2}{*}{gFID↓} & \multirow{2}{*}{$H_\mathcal{C}$} \\ \cmidrule(r){1-2}
Encoder              & Decoder             &                        &                       &                                  \\ \midrule
\multicolumn{2}{@{}l|}{\textcolor{gray}{(Setting \#8 w/o alignment)}} & 2.41 & 11.59        & 7.99                             \\
\checkmark           &                     & 2.19                   & 7.56                  & 9.49                             \\
                     & \checkmark          & 1.91                   & 7.76                  & 10.31                            \\
\checkmark           & \checkmark          & 2.11                   & \underline{4.56}      & 8.79                             \\ \bottomrule
\end{tabular}%
}
\vspace{-6pt}
\end{table}

By conducting the ablations above, we obtain the optimal designs for ResTok which are also used in the main experiments. We also conclude the following key findings: (1) Codebook entropy $H_\mathcal{C}$ matters. Though codebook utilization reflects the ceiling of reconstruction, $H_\mathcal{C}$ is a more important indicator for generation. A higher value of $H_\mathcal{C}$ means that the latent distribution is more dispersed, which is harder for a generator to model, yielding a poorer gFID. (2) Hierarchies significantly enhance representation capacities, but the tokenizer is still suffering from a high value of $H_\mathcal{C}$ and poor generation performance. (3) Residuals guide the tokenizer to add compensatory information around the latent centroids, avoiding dispersing the latent distributions.

\noindent \textbf{Representation Alignment.} As a semantic guidance, the designs of representation alignment affect the convergence. We ablate the alignment positions on setting \#8, resulting in \cref{tab:abl_align}. It demonstrates that aligning representations solely on either the encoder or decoder side is suboptimal, an aspect unexplored in prior work~\citep{yao2025vavae,liu2025detailflow,xiong2025gigatok,zheng2025vfmtok}. Alignments should be applied to the encoder to guide feature extraction, and to the decoder to preserve semantics in the quantization bottleneck, both contributing to improved performance.

\begin{table}[t]
\centering
\caption{Ablation study on the hierarchical AR generator.}
\vspace{-4pt}
\label{tab:abl_har}
{\small
\begin{tabular}{@{}l|lllll@{}}
\toprule
AR Type                    & \#Steps & gFID↓ & IS↑   & Pre.↑ & Rec.↑ \\ \midrule
Vanilla AR                 & 128     & 4.56  & 142.2 & 0.79  & 0.56  \\
Hiera. AR                  &         &       &       &       &       \\
\hspace{1ex} w/o NTP group & 8       & 5.85  & 130.4 & 0.78  & 0.55  \\
\hspace{1ex} w/ NTP group  & 9       & 5.53  & 130.9 & 0.78  & 0.56  \\ \bottomrule
\end{tabular}%
}
\vspace{-6pt}
\end{table}

\begin{figure}[t]
\centering
\includegraphics[width=0.9\linewidth]{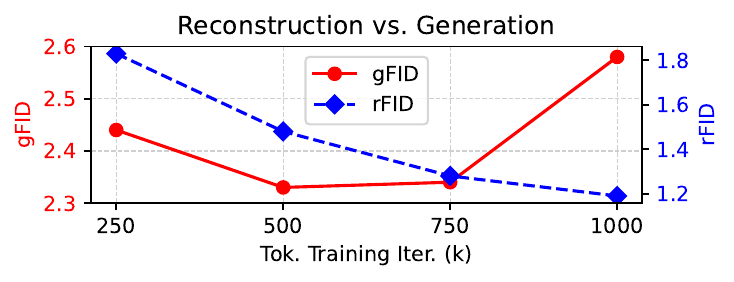}
\vspace{-4pt}
\caption{Reconstruction and generation performance versus tokenizer training iterations.}
\label{fig:recon_vs_gen}
\vspace{-6pt}
\end{figure}

\noindent \textbf{HAR Generation.} We also compare the hierarchical prediction with vanilla AR. As shown in \cref{tab:abl_har}, when switching from vanilla AR to HAR generation, the gFID metric shows an acceptable degradation while the number of sampling steps is dramatically reduced from 128 to 8 or 9. Moreover, introducing a group of NTP tokens (\ie, vanilla AR Gen. in \cref{fig:har}) further reduces sampling errors and improves generation performance.

\noindent \textbf{Recon. vs. Gen.} As the tokenizer trains longer, it may learn overly complex latent patterns that enhance reconstruction but hinder AR modeling. To find a suitable trade-off, we ablate tokenizer training at $\{\text{250k, 500k, 750k, 1M}\}$ iterations, each paired with a fully trained HAR generator. As shown in \cref{fig:recon_vs_gen}, rFID improves steadily with training, whereas gFID reaches its optimum at around 750k steps, after which generation quality degrades. We therefore adopt the 750k tokenizer checkpoint for all main experiments.

\section{Conclusion}

This paper introduced \textbf{Res}idual \textbf{Tok}enizer (\textbf{ResTok}), a 1D visual tokenizer that brings the hierarchical and residual nature of visual representations back to ViT-based tokenizers for autoregressive image generation. Unlike existing isotropic tokenizers that query visual features along only depth, ResTok progressively merges image tokens and accumulates semantic residuals across levels. This hierarchical structure enables latent tokens to organize in a coarse-to-fine manner, achieving natural alignment between image and latent hierarchies without hand-crafted constraints. Extensive experiments verify the effectiveness of hierarchical residuals and implicit alignments in enhancing both reconstruction and generation efficiencies. Future work will further enhance fidelity and explore extension to unified understanding and generation models.

{
    \small
    \bibliographystyle{ieeenat_fullname}
    \bibliography{main}
}

\input{appendix}

\end{document}

%% file: preamble.tex
\usepackage{enumitem}
\usepackage{xspace}
\usepackage{graphicx}
\usepackage{times}
\usepackage{wrapfig}
\usepackage{caption}
\usepackage{subcaption}
\usepackage{booktabs}
\usepackage{amsmath}
\usepackage{amssymb}
\usepackage{bm}
\usepackage{algorithm,algpseudocode}
\usepackage{multirow}
\usepackage{amssymb}
\usepackage{arydshln}
\usepackage{colortbl}

\AtEndPreamble{
    \usepackage[capitalize]{cleveref}
    \crefname{section}{Sec.}{Secs.}
    \Crefname{section}{Section}{Sections}
    \Crefname{table}{Table}{Tables}
    \crefname{table}{Tab.}{Tabs.}
}

\makeatletter
\def\adl@drawiv#1#2#3{%
        \hskip.0\tabcolsep
        \xleaders#3{#2.5\@tempdimb #1{1}#2.5\@tempdimb}%
                #2\z@ plus1fil minus1fil\relax
        \hskip.0\tabcolsep}
\newcommand{\cdashlinelr}[1]{%
  \noalign{\vskip\aboverulesep
           \global\let\@dashdrawstore\adl@draw
           \global\let\adl@draw\adl@drawiv}
  \cdashline{#1}
  \noalign{\global\let\adl@draw\@dashdrawstore
           \vskip\belowrulesep}}
\makeatother

\newlength\savewidth\newcommand\shline{\noalign{\global\savewidth\arrayrulewidth
  \global\arrayrulewidth 1pt}\hline\noalign{\global\arrayrulewidth\savewidth}}
\newcommand{\tablestyle}[2]{\setlength{\tabcolsep}{#1}\renewcommand{\arraystretch}{#2}\centering\footnotesize}

\newcolumntype{x}[1]{>{\centering\arraybackslash}p{#1pt}}
\newcolumntype{y}[1]{>{\raggedright\arraybackslash}p{#1pt}}
\newcolumntype{z}[1]{>{\raggedleft\arraybackslash}p{#1pt}}

\definecolor{lightgray}{gray}{0.9}

%% file: appendix.tex
\clearpage
\appendix

\section*{Appendix}

\begin{figure}[ht]
\centering
\subfloat[Tokenizer mask.]{\includegraphics[width=0.47\linewidth]{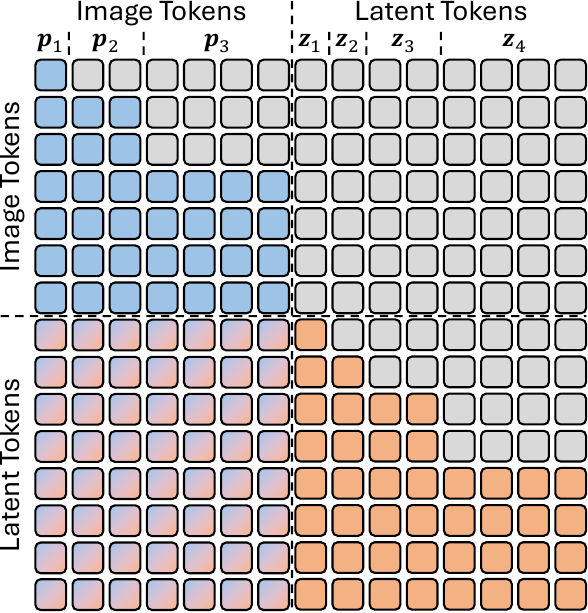}\label{subfig:tok_mask}}\hspace{4pt}
\subfloat[Generator mask.]{\includegraphics[width=0.49\linewidth]{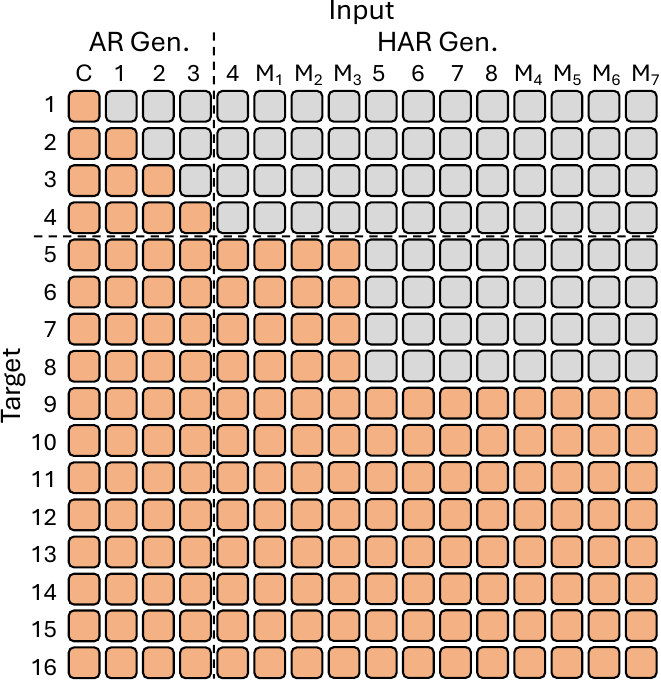}\label{subfig:gen_mask}}
\caption{Implementations of attention masks in the tokenizer and the generator. The tokenizer mask is illustrated using 3 image-token scales and 4 latent hierarchies as an example, while the generator mask is shown with 4 vanilla AR tokens and 2 groups of HAR tokens.}
\label{fig:attn_mask}
\end{figure}

\begin{figure}[ht]
\centering
\subfloat[Encoder RoPE.]{\includegraphics[scale=0.47]{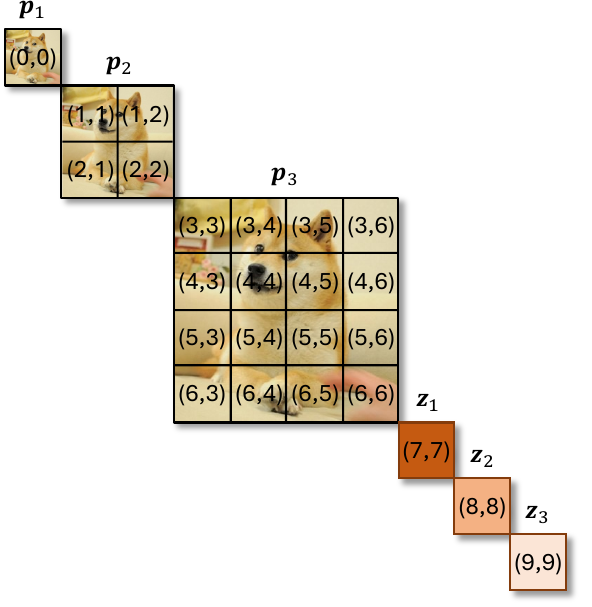}}
\subfloat[Decoder RoPE.]{\includegraphics[scale=0.47]{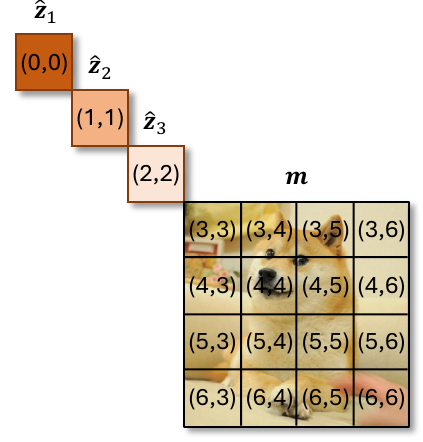}}
\caption{Implementations of 2D RoPE in ResTok, illustrated using 3 image-token scales and 3 latent tokens as an example.}
\label{fig:rope}
\end{figure}

\section{More Implementation Details}
\label{appx:impl}

\subsection{Architecture}

For the CNN encoder and decoder, we adopt exact the same configuration of MaskGIT's encoder and decoder~\citep{chang2022maskgit}. For the ViT encoder and decoder, we develop them upon TiTok-L's architecture~\citep{yu2024titok}, each comprising 24 transformer layers, 1024 dimensions and 16 heads. To bridge the dimension of the CNN encoder/decoder and the ViT encoder/decoder, an additional linear layer is applied between them. We apply encoder attention masks as shown in \cref{subfig:tok_mask} to enforce the causality of encoding process. Additionally, we replace learnable positional embeddings in the original TiTok with a modified 2D version of M-RoPE~\citep{wang2024qwen2vl}, which takes 1D latent tokens as ``text'' and 2D image tokens as ``image'' as shown in \cref{fig:rope}. Specifically, the positional IDs of image tokens from multiple hierarchies are concatenated sequentially, together with those of the text tokens. In the encoder, M-RoPE is applied in the order of coarse-to-fine 2D image tokens, followed by the 1D latent tokens. In the decoder, the sequence begins with the 1D latent tokens, which are then followed by the 2D masked image tokens. The residual 1D latent token initialization and the residual merging process proposed in \cref{fig:pipeline} can be formally represented as \cref{alg:res_1d} and \cref{alg:res_merge}, respectively. For the generator, we apply the attention mask as shown in \cref{subfig:gen_mask} to enable next-hierarchy prediction.

\begin{algorithm}[t]
    \caption{Residual 1D latent token initialization}\label{alg:res_1d}
    \begin{algorithmic}[1]
        \Require{image tokens $\bm{p}^{(0)}$, hierarchical levels $L$.}
        \State $h = 1$, $w = 1$
        \State $\bm{z}^{(0)}_1 = \text{Pool}_{h \times w}(\bm{p}^{(0)})$
        \For{$l = 2, 3, \ldots, L$}
            \State $\bm{p}^{(0)} = \bm{p}^{(0)} - \text{Upsample}(\bm{z}^{(0)}_{l-1})$
            \State $\bm{z}^{(0)}_l = \text{Pool}_{h \times w}(\bm{p}^{(0)})$
            \State $\bm{z}^{(0)}_{1:l} = \text{Concat}(\bm{z}^{(0)}_{1:l-1}, \bm{z}^{(0)}_l)$
            \If{$l~\%~2 = 0$}
                \State $w = w \cdot 2$
            \Else
                \State $h = h \cdot 2$
            \EndIf
        \EndFor
        \State \Return latent tokens $\bm{z}^{(0)}_{1:L}$
    \end{algorithmic}
\end{algorithm}

\begin{algorithm}[t]
    \caption{Residual merging process}\label{alg:res_merge}
    \begin{algorithmic}[1]
        \Require{image tokens $\bm{p}^{(n)}_{\ge s}$, latent tokens $\bm{z}^{(n)}_{1:L}$.}
        \State $\{ \bm{p}^{(n)}_{\ge s}, \bm{z}^{(n)}_{1:L} \} = \text{Attention}(\{ \bm{p}^{(n)}_{\ge s}, \bm{z}^{(n)}_{1:L} \})$
        \State $\bm{p}^{(n)}_{s-1} = \text{Merge}(\bm{p}^{(n)}_{s})$
        \State $\bm{p}^{(n)}_{s} = \bm{p}^{(n)}_{s} - \text{Upsample}(\bm{p}^{(n)}_{s-1})$
        \State $\{ \bm{p}^{(n+1)}_{s-1}, \bm{p}^{(n+1)}_{\ge s}, \bm{z}^{(n+1)}_{1:L} \} = \text{MLP}(\{ \bm{p}^{(n)}_{s-1}, \bm{p}^{(n)}_{\ge s}, \bm{z}^{(n)}_{1:L} \})$
        \State $\bm{p}^{(n+1)}_{\ge s-1} = \text{Concat}(\bm{p}^{(n+1)}_{s-1}, \bm{p}^{(n+1)}_{\ge s})$
        \State \Return image tokens $\bm{p}^{(n+1)}_{\ge s-1}$ and latent tokens $\bm{z}^{(n+1)}_{1:L}$
    \end{algorithmic}
\end{algorithm}

\begin{table*}[t]
\centering
\caption{Classifier-free guidance (CFG) configurations used for different tokenizer checkpoints. For ``Step'' schedules, guidance is activated at the specified ``CFG Start Ratio'' of the sampling trajectory with a fixed ``Max. CFG Value''. For ``Linear'' schedules, the CFG value increases linearly from 1.0 to the ``Max. CFG Value'' over the full sampling process. During sampling, we first apply Top-\textit{K} filtering followed by Top-\textit{P} (nucleus) filtering. Setting the value of \textit{K} or \textit{P} to 0 indicates bypassing Top-\textit{K} or Top-\textit{P} filtering.}
\label{tab:cfg}
{\small
\begin{tabular}{@{}llllll|llll@{}}
\toprule
Ckpt. & Schedule & CFG Start Ratio & Max. CFG Value & Top-\textit{K} & Top-\textit{P} & gFID↓ & IS↑   & Pre.↑ & Rec.↑ \\ \midrule
250K  & Step     & 50\%            & 4.50           & 0              & 0.99           & 2.44  & 230.7 & 0.79  & 0.59  \\
500K  & Step     & 25\%            & 4.50           & 0              & 0.99           & 2.33  & 249.1 & 0.78  & 0.60  \\
750K  & Step     & 25\%            & 3.75           & 0              & 0.95           & 2.34  & 257.8 & 0.79  & 0.60  \\
1M    & Linear   & N/A             & 4.00           & 0              & 0.95           & 2.58  & 252.3 & 0.78  & 0.61  \\ \bottomrule
\end{tabular}%
}
\end{table*}

\subsection{Training}

\begin{table}[t]
\centering
\tablestyle{6pt}{1.02}
\caption{Training settings of ResTok.}
\label{tab:training_tok}
\small
\begin{tabular}{y{96}|y{96}}
config & value \\
\shline
optimizer & AdamW~\citep{loshchilov2019adamw} \\
base learning rate & 1\textit{e}-4 \\
weight decay & 1\textit{e}-4 \\
optimizer momentum & $\beta_1, \beta_2{=}0.9, 0.95$ \\
batch size & 256 \\
learning rate schedule & cosine decay \\
minimal learning rate & 1\textit{e}-5 \\
training epochs & 200 \\
linear warmup epochs & 1 \\
augmentation & RandomResizedCrop \\
ema decay & 0.9999
\end{tabular}
\end{table}

\begin{table}[t]
\centering
\tablestyle{6pt}{1.02}
\caption{Training settings of LlamaGen-L.}
\label{tab:training_gen}
\small
\begin{tabular}{y{96}|y{96}}
config & value \\
\shline
optimizer & AdamW~\citep{loshchilov2019adamw} \\
base learning rate & 1\textit{e}-4 \\
weight decay & 0.05 \\
optimizer momentum & $\beta_1, \beta_2{=}0.9, 0.95$ \\
batch size & 256 \\
learning rate schedule & cosine decay \\
minimal learning rate & 1\textit{e}-5 \\
training epochs & 300 \\
linear warmup epochs & 1 \\
augmentation & ResizedCrop \\
ema decay & 0.9999
\end{tabular}
\end{table}

Our training configurations of ResTok and LlamaGen-L~\citep{sun2024llamagen} are listed in \cref{tab:training_tok,tab:training_gen}. Both the tokenizer and the generator are trained from scratch on the ImageNet-1K training set~\citep{imagenet1k}, consisting of 1,281,167 images across 1,000 object classes. When training ResTok, images are first randomly resized with a factor between $[0.8,1.0]$, and then cropped to 256$\times$256 at a random position. To prepare the training data for the generator, we use the same scripts and data augmentations to extract quantized codes as LlamaGen~\citep{sun2024llamagen}.
We set $\lambda_\text{enc} = \lambda_\text{dec} = \lambda_\text{vf} = \lambda_\text{mse} = \lambda_\text{percp} = 1.0$ and $\lambda_\text{gan} = 0.5$ in \cref{eq:loss_vf,eq:loss_total}.

We apply nested token dropout~\citep{miwa2025onedpiece,bachmann2025flextok,li2025imagefolder,liu2025detailflow} during training. The keeping probabilities for each token length are listed in \cref{tab:drop}, with a minimum of 4 tokens preserved. In our setting, there is an 80\% chance that no dropout is applied, while the dropout probability for shorter token lengths decreases exponentially as the target length decreases.

\subsection{Evaluation}

\begin{table}[t]
\centering
\caption{Keeping probabilities of nested token dropout.}
\label{tab:drop}
{\small
\resizebox{\linewidth}{!}{%
\begin{tabular}{@{}l|llllll@{}}
\toprule
\#Tokens    & 128     & 64      & 32     & 16     & 8      & 4      \\ \midrule
Probability & 80.00\% & 10.32\% & 5.16\% & 2.58\% & 1.29\% & 0.65\% \\ \bottomrule
\end{tabular}%
}
}
\end{table}

\begin{table}[t]
\centering
\caption{Additional results of AR generation on ResTok.}
\label{tab:addi_ar}
{\small
\begin{tabular}{@{}l|lllll@{}}
\toprule
AR Type    & \#Steps & gFID↓ & IS↑   & Pre.↑ & Rec.↑ \\ \midrule
HAR        & 9       & 2.34  & 257.8 & 0.79  & 0.60  \\
Vanilla AR & 128     & 2.18  & 259.1 & 0.79  & 0.62  \\ \bottomrule
\end{tabular}%
}
\end{table}

To evaluate ResTok's reconstruction ability, we utilize the same protocol as TiTok~\citep{yu2024titok}. To obtain the metrics of generation performance, we use the same scripts as GigaTok~\citep{xiong2025gigatok} to generate images and calculate gFID, IS, Precision and Recall. Specifically, we search for the best CFG~\citep{ho2021cfg} schedules of each HAR generator corresponding to each checkpoint of ResTok in \cref{fig:recon_vs_gen}, which are listed in \cref{tab:cfg}. The best trade-off (\ie, the 750K step checkpoint) is selected as the final model. Ablations in \cref{sec:ablation} which take the 150K step checkpoint of the tokenizer and the 250K step checkpoint of the generator, do not enable CFG for evaluation.

To quantify the distributional uniformity of codebook usage, we compute the empirical entropy of the selected codebook entries.
Let the codebook $\mathcal{C}$ contain $K$ entries. For each entry $i \in \{ 1, \ldots, K \}$, let $c_i$ denote the number of times it is selected during evaluation, the empirical probability of selecting entry $i$ is
\begin{equation}
p_i = \frac{c_i}{\sum^K_{j=1} c_j}.
\end{equation}
The codebook entropy $H_\mathcal{C}$ is then defined as the standard Shannon entropy (measured in bits)
\begin{equation}
H_\mathcal{C} = -\sum^K_{i=1} p_i \log_2 (p_i + \epsilon),
\end{equation}
where a small constant $\epsilon$ is added for numerical stability. We set $\epsilon = 1 \times 10^{-8}$ as TiTok~\citep{yu2024titok} does. A higher value of $H_\mathcal{C}$ indicates more uniform codebook usage, while lower entropy suggests concentration on a small subset of entries.

\section{Additional Results}

In addition to the HAR version reported in \cref{tab:main_cmp}, we also train a vanilla AR variant to evaluate the upper bound of AR generation performance on ResTok. The results are presented in \cref{tab:addi_ar}. The vanilla AR model uses a \textit{step} CFG schedule, where CFG is activated after sampling the first 4 tokens with a fixed value of 4.5. Compared with HAR, which requires only 9 sampling steps, vanilla AR reduces gFID from 2.34 to 2.18 but incurs more than a 10$\times$ increase in sampling steps, demonstrating the effectiveness of our proposed approach.

\section{Licenses for Released Assets}
\label{sec:licenses}

This work uses the listed projects in \cref{tab:licenses} released under their licenses. We strictly adhered to their license requirements; the original projects' copyright notices and license texts can be found in their official repositories.

\begin{table}[t]
\centering
\caption{Licenses for released assets}
\label{tab:licenses}
{\small
\begin{tabular}{@{}ll@{}}
\toprule
\textbf{Asset} & \textbf{License} \\ \midrule
TiTok~\citep{yu2024titok} & Apache-2.0 license \\ \midrule
LlamaGen~\citep{sun2024llamagen} & MIT license \\ \midrule
GigaTok~\citep{xiong2025gigatok} & MIT license \\ \midrule
VA-VAE~\citep{yao2025vavae} & MIT license \\ \midrule
DINOv3~\citep{simeoni2025dinov3} & DINOv3 License \\ \midrule
ImageNet-1K~\cite{imagenet1k} & Custom (research-only, non-commercial) \\ \bottomrule
\end{tabular}
}
\end{table}

%% file: main.bib
@String(CVPR= {IEEE Conf. Comput. Vis. Pattern Recog.})

@String(ICCV= {Int. Conf. Comput. Vis.})

@String(CVPR  = {CVPR})

@String(ICCV  = {ICCV})

@article{
hinton2006ae,
author = {Hinton, Geoffrey E. and Salakhutdinov, Ruslan R.},
title = {Reducing the Dimensionality of Data with Neural Networks},
journal = {Science},
volume = {313},
number = {5786},
pages = {504-507},
year = {2006},
doi = {10.1126/science.1127647},
URL = {https://www.science.org/doi/abs/10.1126/science.1127647},
eprint = {https://www.science.org/doi/pdf/10.1126/science.1127647}}

@inproceedings{ian2014gan,
    author = {Goodfellow, Ian and Pouget-Abadie, Jean and Mirza, Mehdi and Xu, Bing and Warde-Farley, David and Ozair, Sherjil and Courville, Aaron and Bengio, Yoshua},
    booktitle = {Advances in Neural Information Processing Systems},
    editor = {Z. Ghahramani and M. Welling and C. Cortes and N. Lawrence and K.Q. Weinberger},
    pages = {},
    publisher = {Curran Associates, Inc.},
    title = {Generative Adversarial Nets},
    url = {https://proceedings.neurips.cc/paper_files/paper/2014/file/5ca3e9b122f61f8f06494c97b1afccf3-Paper.pdf},
    volume = {27},
    year = {2014}
}

@inproceedings{
kingma2014vae,
title={Auto-Encoding Variational Bayes},
author={Kingma, Diederik P. and Welling, Max},
booktitle={International Conference on Learning Representations},
year={2014},
url={https://openreview.net/forum?id=33X9fd2-9FyZd}
}

@InProceedings{gregor2015draw,
  title = 	 {DRAW: A Recurrent Neural Network For Image Generation},
  author = 	 {Gregor, Karol and Danihelka, Ivo and Graves, Alex and Rezende, Danilo and Wierstra, Daan},
  booktitle = 	 {Proceedings of the 32nd International Conference on Machine Learning},
  pages = 	 {1462--1471},
  year = 	 {2015},
  editor = 	 {Bach, Francis and Blei, David},
  volume = 	 {37},
  series = 	 {Proceedings of Machine Learning Research},
  address = 	 {Lille, France},
  month = 	 {07--09 Jul},
  publisher =    {PMLR},
  url = 	 {https://proceedings.mlr.press/v37/gregor15.html}
}

@inproceedings{kingma2016iaf,
 author = {Kingma, Diederik P. and Salimans, Tim and Jozefowicz, Rafal and Chen, Xi and Sutskever, Ilya and Welling, Max},
 booktitle = {Advances in Neural Information Processing Systems},
 editor = {D. Lee and M. Sugiyama and U. Luxburg and I. Guyon and R. Garnett},
 pages = {},
 publisher = {Curran Associates, Inc.},
 title = {Improved Variational Inference with Inverse Autoregressive Flow},
 url = {https://proceedings.neurips.cc/paper_files/paper/2016/file/ddeebdeefdb7e7e7a697e1c3e3d8ef54-Paper.pdf},
 volume = {29},
 year = {2016}
}

@inproceedings{sonderby2016ladder,
 author = {S{\o}nderby, Casper Kaae and Raiko, Tapani and Maal{\o}e, Lars and S{\o}nderby, S{\o}ren Kaae and Winther, Ole},
 booktitle = {Advances in Neural Information Processing Systems},
 editor = {D. Lee and M. Sugiyama and U. Luxburg and I. Guyon and R. Garnett},
 pages = {},
 publisher = {Curran Associates, Inc.},
 title = {Ladder Variational Autoencoders},
 url = {https://proceedings.neurips.cc/paper_files/paper/2016/file/6ae07dcb33ec3b7c814df797cbda0f87-Paper.pdf},
 volume = {29},
 year = {2016}
}

@inproceedings{oord2017vqvae,
 author = {van den Oord, Aäron and Vinyals, Oriol and kavukcuoglu, koray},
 booktitle = {Advances in Neural Information Processing Systems},
 editor = {I. Guyon and U. Von Luxburg and S. Bengio and H. Wallach and R. Fergus and S. Vishwanathan and R. Garnett},
 pages = {},
 publisher = {Curran Associates, Inc.},
 title = {Neural Discrete Representation Learning},
 url = {https://proceedings.neurips.cc/paper_files/paper/2017/file/7a98af17e63a0ac09ce2e96d03992fbc-Paper.pdf},
 volume = {30},
 year = {2017}
}

@InProceedings{oord2016pixelrnn,
  title = 	 {Pixel Recurrent Neural Networks},
  author = 	 {van den Oord, Aäron and Kalchbrenner, Nal and Kavukcuoglu, Koray},
  booktitle = 	 {Proceedings of The 33rd International Conference on Machine Learning},
  pages = 	 {1747--1756},
  year = 	 {2016},
  editor = 	 {Balcan, Maria Florina and Weinberger, Kilian Q.},
  volume = 	 {48},
  series = 	 {Proceedings of Machine Learning Research},
  address = 	 {New York, New York, USA},
  month = 	 {20--22 Jun},
  publisher =    {PMLR},
  url = 	 {https://proceedings.mlr.press/v48/oord16.html}
}

@inproceedings{oord2016pixelcnn,
 author = {van den Oord, Aäron and Kalchbrenner, Nal and Espeholt, Lasse and kavukcuoglu, koray and Vinyals, Oriol and Graves, Alex},
 booktitle = {Advances in Neural Information Processing Systems},
 editor = {D. Lee and M. Sugiyama and U. Luxburg and I. Guyon and R. Garnett},
 pages = {},
 publisher = {Curran Associates, Inc.},
 title = {Conditional Image Generation with PixelCNN Decoders},
 url = {https://proceedings.neurips.cc/paper_files/paper/2016/file/b1301141feffabac455e1f90a7de2054-Paper.pdf},
 volume = {29},
 year = {2016}
}

@inproceedings{vaswani2017transformer,
 author = {Vaswani, Ashish and Shazeer, Noam and Parmar, Niki and Uszkoreit, Jakob and Jones, Llion and Gomez, Aidan N and Kaiser, {\L}ukasz and Polosukhin, Illia},
 booktitle = {Advances in Neural Information Processing Systems},
 editor = {I. Guyon and U. Von Luxburg and S. Bengio and H. Wallach and R. Fergus and S. Vishwanathan and R. Garnett},
 pages = {},
 publisher = {Curran Associates, Inc.},
 title = {Attention is All you Need},
 url = {https://proceedings.neurips.cc/paper_files/paper/2017/file/3f5ee243547dee91fbd053c1c4a845aa-Paper.pdf},
 volume = {30},
 year = {2017}
}

@inproceedings{
dosovitskiy2021vit,
title={An Image is Worth 16x16 Words: Transformers for Image Recognition at Scale},
author={Alexey Dosovitskiy and Lucas Beyer and Alexander Kolesnikov and Dirk Weissenborn and Xiaohua Zhai and Thomas Unterthiner and Mostafa Dehghani and Matthias Minderer and Georg Heigold and Sylvain Gelly and Jakob Uszkoreit and Neil Houlsby},
booktitle={International Conference on Learning Representations},
year={2021},
url={https://openreview.net/forum?id=YicbFdNTTy}
}

@InProceedings{chang2022maskgit,
    author    = {Chang, Huiwen and Zhang, Han and Jiang, Lu and Liu, Ce and Freeman, William T.},
    title     = {{M}ask{GIT}: Masked Generative Image Transformer},
    booktitle = {Proceedings of the IEEE/CVF Conference on Computer Vision and Pattern Recognition (CVPR)},
    month     = {June},
    year      = {2022},
    pages     = {11315-11325}
}

@inproceedings{yu2024titok,
 author = {Yu, Qihang and Weber, Mark and Deng, Xueqing and Shen, Xiaohui and Cremers, Daniel and Chen, Liang-Chieh},
 booktitle = {Advances in Neural Information Processing Systems},
 editor = {A. Globerson and L. Mackey and D. Belgrave and A. Fan and U. Paquet and J. Tomczak and C. Zhang},
 pages = {128940--128966},
 publisher = {Curran Associates, Inc.},
 title = {An Image is Worth 32 Tokens for Reconstruction and Generation},
 url = {https://proceedings.neurips.cc/paper_files/paper/2024/file/e91bf7dfba0477554994c6d64833e9d8-Paper-Conference.pdf},
 volume = {37},
 year = {2024}
}

@InProceedings{bachmann2025flextok,
  title = 	 {{F}lex{T}ok: Resampling Images into 1{D} Token Sequences of Flexible Length},
  author =       {Bachmann, Roman and Allardice, Jesse and Mizrahi, David and Fini, Enrico and Kar, O\u{g}uzhan Fatih and Amirloo, Elmira and El-Nouby, Alaaeldin and Zamir, Amir and Dehghan, Afshin},
  booktitle = 	 {Proceedings of the 42nd International Conference on Machine Learning},
  pages = 	 {2241--2292},
  year = 	 {2025},
  editor = 	 {Singh, Aarti and Fazel, Maryam and Hsu, Daniel and Lacoste-Julien, Simon and Berkenkamp, Felix and Maharaj, Tegan and Wagstaff, Kiri and Zhu, Jerry},
  volume = 	 {267},
  series = 	 {Proceedings of Machine Learning Research},
  month = 	 {13--19 Jul},
  publisher =    {PMLR},
  url = 	 {https://proceedings.mlr.press/v267/bachmann25a.html}
}

@inproceedings{
yu2025repa,
title={Representation Alignment for Generation: Training Diffusion Transformers Is Easier Than You Think},
author={Sihyun Yu and Sangkyung Kwak and Huiwon Jang and Jongheon Jeong and Jonathan Huang and Jinwoo Shin and Saining Xie},
booktitle={The Thirteenth International Conference on Learning Representations},
year={2025},
url={https://openreview.net/forum?id=DJSZGGZYVi}
}

@InProceedings{rombach2022ldm,
    author    = {Rombach, Robin and Blattmann, Andreas and Lorenz, Dominik and Esser, Patrick and Ommer, Bj\"orn},
    title     = {High-Resolution Image Synthesis With Latent Diffusion Models},
    booktitle = {Proceedings of the IEEE/CVF Conference on Computer Vision and Pattern Recognition (CVPR)},
    month     = {June},
    year      = {2022},
    pages     = {10684-10695}
}

@inproceedings{
li2025imagefolder,
title={{I}mage{F}older: Autoregressive Image Generation with Folded Tokens},
author={Xiang Li and Kai Qiu and Hao Chen and Jason Kuen and Jiuxiang Gu and Bhiksha Raj and Zhe Lin},
booktitle={The Thirteenth International Conference on Learning Representations},
year={2025},
url={https://openreview.net/forum?id=QE1LFzXQPL}
}

@inproceedings{xiong2025gigatok,
    author    = {Xiong, Tianwei and Liew, Jun Hao and Huang, Zilong and Feng, Jiashi and Liu, Xihui},
    title     = {{G}iga{T}ok: Scaling Visual Tokenizers to 3 Billion Parameters for Autoregressive Image Generation},
    booktitle = {Proceedings of the IEEE/CVF International Conference on Computer Vision (ICCV)},
    month     = {October},
    year      = {2025},
    pages     = {18770-18780}
}

@inproceedings{huang2025spectralar,
    author    = {Huang, Yuanhui and Chen, Weiliang and Zheng, Wenzhao and Duan, Yueqi and Zhou, Jie and Lu, Jiwen},
    title     = {SpectralAR: Spectral Autoregressive Visual Generation},
    booktitle = {Proceedings of the IEEE/CVF International Conference on Computer Vision (ICCV)},
    month     = {October},
    year      = {2025},
    pages     = {15842-15852}
}

@article{liu2025detailflow,
  title={DetailFlow: 1D Coarse-to-Fine Autoregressive Image Generation via Next-Detail Prediction},
  author={Liu, Yiheng and Qu, Liao and Zhang, Huichao and Wang, Xu and Jiang, Yi and Gao, Yiming and Ye, Hu and Li, Xian and Wang, Shuai and Du, Daniel K and others},
  journal={arXiv preprint arXiv:2505.21473},
  year={2025}
}

@InProceedings{wang2025par,
    author    = {Wang, Yuqing and Ren, Shuhuai and Lin, Zhijie and Han, Yujin and Guo, Haoyuan and Yang, Zhenheng and Zou, Difan and Feng, Jiashi and Liu, Xihui},
    title     = {Parallelized Autoregressive Visual Generation},
    booktitle = {Proceedings of the IEEE/CVF Conference on Computer Vision and Pattern Recognition (CVPR)},
    month     = {June},
    year      = {2025},
    pages     = {12955-12965}
}

@article{sun2024llamagen,
  title={Autoregressive Model Beats Diffusion: Llama for Scalable Image Generation},
  author={Sun, Peize and Jiang, Yi and Chen, Shoufa and Zhang, Shilong and Peng, Bingyue and Luo, Ping and Yuan, Zehuan},
  journal={arXiv preprint arXiv:2406.06525},
  year={2024}
}

@inproceedings{tian2024var,
 author = {Tian, Keyu and Jiang, Yi and Yuan, Zehuan and Peng, Bingyue and Wang, Liwei},
 booktitle = {Advances in Neural Information Processing Systems},
 editor = {A. Globerson and L. Mackey and D. Belgrave and A. Fan and U. Paquet and J. Tomczak and C. Zhang},
 pages = {84839--84865},
 publisher = {Curran Associates, Inc.},
 title = {Visual Autoregressive Modeling: Scalable Image Generation via Next-Scale Prediction},
 url = {https://proceedings.neurips.cc/paper_files/paper/2024/file/9a24e284b187f662681440ba15c416fb-Paper-Conference.pdf},
 volume = {37},
 year = {2024}
}

@inproceedings{
ren2025flowar,
title={Flow{AR}: Scale-wise Autoregressive Image Generation Meets Flow Matching},
author={Sucheng Ren and Qihang Yu and Ju He and Xiaohui Shen and Alan Yuille and Liang-Chieh Chen},
booktitle={Forty-second International Conference on Machine Learning},
year={2025},
url={https://openreview.net/forum?id=JfLgvNe1tj}
}

@inproceedings{li2024mar,
 author = {Li, Tianhong and Tian, Yonglong and Li, He and Deng, Mingyang and He, Kaiming},
 booktitle = {Advances in Neural Information Processing Systems},
 editor = {A. Globerson and L. Mackey and D. Belgrave and A. Fan and U. Paquet and J. Tomczak and C. Zhang},
 pages = {56424--56445},
 publisher = {Curran Associates, Inc.},
 title = {Autoregressive Image Generation without Vector Quantization},
 url = {https://proceedings.neurips.cc/paper_files/paper/2024/file/66e226469f20625aaebddbe47f0ca997-Paper-Conference.pdf},
 volume = {37},
 year = {2024}
}

@InProceedings{yao2025vavae,
    author    = {Yao, Jingfeng and Yang, Bin and Wang, Xinggang},
    title     = {Reconstruction vs. Generation: Taming Optimization Dilemma in Latent Diffusion Models},
    booktitle = {Proceedings of the IEEE/CVF Conference on Computer Vision and Pattern Recognition (CVPR)},
    month     = {June},
    year      = {2025},
    pages     = {15703-15712}
}

@InProceedings{peebles2023dit,
    author    = {Peebles, William and Xie, Saining},
    title     = {Scalable Diffusion Models with Transformers},
    booktitle = {Proceedings of the IEEE/CVF International Conference on Computer Vision (ICCV)},
    month     = {October},
    year      = {2023},
    pages     = {4195-4205}
}

@inproceedings{shi2025ibq,
  author={Shi, Fengyuan and Luo, Zhuoyan and Ge, Yixiao and Yang, Yujiu and Shan, Ying and Wang, Limin},
  title={Scalable Image Tokenization with Index Backpropagation Quantization},
  booktitle={Proceedings of the IEEE/CVF International Conference on Computer Vision (ICCV)},
  month={October},
  year={2025},
  pages={16037-16046}
}

@InProceedings{lee2022rqvae,
    author    = {Lee, Doyup and Kim, Chiheon and Kim, Saehoon and Cho, Minsu and Han, Wook-Shin},
    title     = {Autoregressive Image Generation Using Residual Quantization},
    booktitle = {Proceedings of the IEEE/CVF Conference on Computer Vision and Pattern Recognition (CVPR)},
    month     = {June},
    year      = {2022},
    pages     = {11523-11532}
}

@InProceedings{chen2020igpt,
  title = 	 {Generative Pretraining From Pixels},
  author =       {Chen, Mark and Radford, Alec and Child, Rewon and Wu, Jeffrey and Jun, Heewoo and Luan, David and Sutskever, Ilya},
  booktitle = 	 {Proceedings of the 37th International Conference on Machine Learning},
  pages = 	 {1691--1703},
  year = 	 {2020},
  editor = 	 {III, Hal Daumé and Singh, Aarti},
  volume = 	 {119},
  series = 	 {Proceedings of Machine Learning Research},
  month = 	 {13--18 Jul},
  publisher =    {PMLR},
  url = 	 {https://proceedings.mlr.press/v119/chen20s.html}
}

@InProceedings{esser2021vqgan,
    author    = {Esser, Patrick and Rombach, Robin and Ommer, Bjorn},
    title     = {Taming Transformers for High-Resolution Image Synthesis},
    booktitle = {Proceedings of the IEEE/CVF Conference on Computer Vision and Pattern Recognition (CVPR)},
    month     = {June},
    year      = {2021},
    pages     = {12873-12883}
}

@inproceedings{
yu2022vitvqgan,
title={Vector-quantized Image Modeling with Improved {VQGAN}},
author={Jiahui Yu and Xin Li and Jing Yu Koh and Han Zhang and Ruoming Pang and James Qin and Alexander Ku and Yuanzhong Xu and Jason Baldridge and Yonghui Wu},
booktitle={International Conference on Learning Representations},
year={2022},
url={https://openreview.net/forum?id=pfNyExj7z2}
}

@INPROCEEDINGS{imagenet1k,
  author={Deng, Jia and Dong, Wei and Socher, Richard and Li, Li-Jia and Kai Li and Li Fei-Fei},
  booktitle={2009 IEEE Conference on Computer Vision and Pattern Recognition}, 
  title={{I}mage{N}et: A large-scale hierarchical image database}, 
  year={2009},
  volume={},
  number={},
  pages={248-255},
  doi={10.1109/CVPR.2009.5206848}}

@InProceedings{he2016resnet,
author = {He, Kaiming and Zhang, Xiangyu and Ren, Shaoqing and Sun, Jian},
title = {Deep Residual Learning for Image Recognition},
booktitle = {Proceedings of the IEEE Conference on Computer Vision and Pattern Recognition (CVPR)},
month = {June},
year = {2016},
pages = {770-778}
}

@InProceedings{liu2021swin,
    author    = {Liu, Ze and Lin, Yutong and Cao, Yue and Hu, Han and Wei, Yixuan and Zhang, Zheng and Lin, Stephen and Guo, Baining},
    title     = {Swin Transformer: Hierarchical Vision Transformer Using Shifted Windows},
    booktitle = {Proceedings of the IEEE/CVF International Conference on Computer Vision (ICCV)},
    month     = {October},
    year      = {2021},
    pages     = {10012-10022}
}

@InProceedings{liu2022convnext,
    author    = {Liu, Zhuang and Mao, Hanzi and Wu, Chao-Yuan and Feichtenhofer, Christoph and Darrell, Trevor and Xie, Saining},
    title     = {A ConvNet for the 2020s},
    booktitle = {Proceedings of the IEEE/CVF Conference on Computer Vision and Pattern Recognition (CVPR)},
    month     = {June},
    year      = {2022},
    pages     = {11976-11986}
}

@inproceedings{
yu2024magvit2,
title={Language Model Beats Diffusion - Tokenizer is key to visual generation},
author={Lijun Yu and Jose Lezama and Nitesh Bharadwaj Gundavarapu and Luca Versari and Kihyuk Sohn and David Minnen and Yong Cheng and Agrim Gupta and Xiuye Gu and Alexander G Hauptmann and Boqing Gong and Ming-Hsuan Yang and Irfan Essa and David A Ross and Lu Jiang},
booktitle={The Twelfth International Conference on Learning Representations},
year={2024},
url={https://openreview.net/forum?id=gzqrANCF4g}
}

@inproceedings{yu2023spae,
 author = {Yu, Lijun and Cheng, Yong and Wang, Zhiruo and Kumar, Vivek and Macherey, Wolfgang and Huang, Yanping and Ross, David and Essa, Irfan and Bisk, Yonatan and Yang, Ming-Hsuan and Murphy, Kevin P and Hauptmann, Alexander and Jiang, Lu},
 booktitle = {Advances in Neural Information Processing Systems},
 editor = {A. Oh and T. Naumann and A. Globerson and K. Saenko and M. Hardt and S. Levine},
 pages = {52692--52704},
 publisher = {Curran Associates, Inc.},
 title = {{SPAE}: Semantic Pyramid AutoEncoder for Multimodal Generation with Frozen LLMs},
 url = {https://proceedings.neurips.cc/paper_files/paper/2023/file/a526cc8f6ffb74bedb6ff313e3fdb450-Paper-Conference.pdf},
 volume = {36},
 year = {2023}
}

@InProceedings{beyer2025highly,
  title = 	 {Highly Compressed Tokenizer Can Generate Without Training},
  author =       {Beyer, Lukas Lao and Li, Tianhong and Chen, Xinlei and Karaman, Sertac and He, Kaiming},
  booktitle = 	 {Proceedings of the 42nd International Conference on Machine Learning},
  pages = 	 {4096--4114},
  year = 	 {2025},
  editor = 	 {Singh, Aarti and Fazel, Maryam and Hsu, Daniel and Lacoste-Julien, Simon and Berkenkamp, Felix and Maharaj, Tegan and Wagstaff, Kiri and Zhu, Jerry},
  volume = 	 {267},
  series = 	 {Proceedings of Machine Learning Research},
  month = 	 {13--19 Jul},
  publisher =    {PMLR},
  url = 	 {https://proceedings.mlr.press/v267/beyer25a.html}
}

@InProceedings{han2025infinity,
    author    = {Han, Jian and Liu, Jinlai and Jiang, Yi and Yan, Bin and Zhang, Yuqi and Yuan, Zehuan and Peng, Bingyue and Liu, Xiaobing},
    title     = {Infinity: Scaling Bitwise AutoRegressive Modeling for High-Resolution Image Synthesis},
    booktitle = {Proceedings of the IEEE/CVF Conference on Computer Vision and Pattern Recognition (CVPR)},
    month     = {June},
    year      = {2025},
    pages     = {15733-15744}
}

@inproceedings{zheng2025vfmtok,
 author = {Anlin Zheng and Xin Wen and Xuanyang Zhang and Chuofan Ma and Tiancai Wang and Gang Yu and Xiangyu Zhang and Xiaojuan Qi},
 booktitle = {Advances in Neural Information Processing Systems},
 title = {Vision Foundation Models as Effective Visual Tokenizers for Autoregressive Generation},
 year = {2025}
}

@inproceedings{
lipman2023flow,
title={Flow Matching for Generative Modeling},
author={Yaron Lipman and Ricky T. Q. Chen and Heli Ben-Hamu and Maximilian Nickel and Matthew Le},
booktitle={The Eleventh International Conference on Learning Representations },
year={2023},
url={https://openreview.net/forum?id=PqvMRDCJT9t}
}

@InProceedings{li2023blip2,
  title = 	 {{BLIP}-2: Bootstrapping Language-Image Pre-training with Frozen Image Encoders and Large Language Models},
  author =       {Li, Junnan and Li, Dongxu and Savarese, Silvio and Hoi, Steven},
  booktitle = 	 {Proceedings of the 40th International Conference on Machine Learning},
  pages = 	 {19730--19742},
  year = 	 {2023},
  editor = 	 {Krause, Andreas and Brunskill, Emma and Cho, Kyunghyun and Engelhardt, Barbara and Sabato, Sivan and Scarlett, Jonathan},
  volume = 	 {202},
  series = 	 {Proceedings of Machine Learning Research},
  month = 	 {23--29 Jul},
  publisher =    {PMLR},
  url = 	 {https://proceedings.mlr.press/v202/li23q.html}
}

@article{simeoni2025dinov3,
  title={{DINO}v3},
  author={Oriane Siméoni and Huy V. Vo and Maximilian Seitzer and Federico Baldassarre and Maxime Oquab and Cijo Jose and Vasil Khalidov and Marc Szafraniec and Seungeun Yi and Michaël Ramamonjisoa and Francisco Massa and Daniel Haziza and Luca Wehrstedt and Jianyuan Wang and Timothée Darcet and Théo Moutakanni and Leonel Sentana and Claire Roberts and Andrea Vedaldi and Jamie Tolan and John Brandt and Camille Couprie and Julien Mairal and Hervé Jégou and Patrick Labatut and Piotr Bojanowski},
  journal={arXiv preprint arXiv:2508.10104},
  year={2025}
}

@article{wang2024qwen2vl,
  title={Qwen2-vl: Enhancing vision-language model's perception of the world at any resolution},
  author={Peng Wang and Shuai Bai and Sinan Tan and Shijie Wang and Zhihao Fan and Jinze Bai and Keqin Chen and Xuejing Liu and Jialin Wang and Wenbin Ge and Yang Fan and Kai Dang and Mengfei Du and Xuancheng Ren and Rui Men and Dayiheng Liu and Chang Zhou and Jingren Zhou and Junyang Lin},
  journal={arXiv preprint arXiv:2409.12191},
  year={2024}
}

@inproceedings{
loshchilov2019adamw,
title={Decoupled Weight Decay Regularization},
author={Ilya Loshchilov and Frank Hutter},
booktitle={International Conference on Learning Representations},
year={2019},
url={https://openreview.net/forum?id=Bkg6RiCqY7},
}

@inproceedings{heusel2017fid,
 author = {Heusel, Martin and Ramsauer, Hubert and Unterthiner, Thomas and Nessler, Bernhard and Hochreiter, Sepp},
 booktitle = {Advances in Neural Information Processing Systems},
 editor = {I. Guyon and U. Von Luxburg and S. Bengio and H. Wallach and R. Fergus and S. Vishwanathan and R. Garnett},
 pages = {},
 publisher = {Curran Associates, Inc.},
 title = {{GAN}s Trained by a Two Time-Scale Update Rule Converge to a Local Nash Equilibrium},
 url = {https://proceedings.neurips.cc/paper_files/paper/2017/file/8a1d694707eb0fefe65871369074926d-Paper.pdf},
 volume = {30},
 year = {2017}
}

@inproceedings{salimans2016is,
 author = {Salimans, Tim and Goodfellow, Ian and Zaremba, Wojciech and Cheung, Vicki and Radford, Alec and Chen, Xi and Chen, Xi},
 booktitle = {Advances in Neural Information Processing Systems},
 editor = {D. Lee and M. Sugiyama and U. Luxburg and I. Guyon and R. Garnett},
 pages = {},
 publisher = {Curran Associates, Inc.},
 title = {Improved Techniques for Training GANs},
 url = {https://proceedings.neurips.cc/paper_files/paper/2016/file/8a3363abe792db2d8761d6403605aeb7-Paper.pdf},
 volume = {29},
 year = {2016}
}

@article{ge2023seed,
  title={Planting a {SEED} of Vision in Large Language Model},
  author={Ge, Yuying and Ge, Yixiao and Zeng, Ziyun and Wang, Xintao and Shan, Ying},
  journal={arXiv preprint arXiv:2307.08041},
  year={2023}
}

@article{miwa2025onedpiece,
  title={{O}ne-{D}-{P}iece: Image tokenizer meets quality-controllable compression},
  author={Miwa, Keita and Sasaki, Kento and Arai, Hidehisa and Takahashi, Tsubasa and Yamaguchi, Yu},
  journal={arXiv preprint arXiv:2501.10064},
  year={2025}
}

@InProceedings{wen2025semanticist,
    author    = {Wen, Xin and Zhao, Bingchen and Elezi, Ismail and Deng, Jiankang and Qi, Xiaojuan},
    title     = {``{P}rincipal Components'' Enable A New Language of Images},
    booktitle = {Proceedings of the IEEE/CVF International Conference on Computer Vision (ICCV)},
    month     = {October},
    year      = {2025},
    pages     = {16641-16651}
}

@InProceedings{zhang2018lpips,
author = {Zhang, Richard and Isola, Phillip and Efros, Alexei A. and Shechtman, Eli and Wang, Oliver},
title = {The Unreasonable Effectiveness of Deep Features as a Perceptual Metric},
booktitle = {Proceedings of the IEEE Conference on Computer Vision and Pattern Recognition (CVPR)},
month = {June},
year = {2018},
pages = {586-595}
}

@inproceedings{
ho2021cfg,
title={Classifier-Free Diffusion Guidance},
author={Jonathan Ho and Tim Salimans},
booktitle={NeurIPS 2021 Workshop on Deep Generative Models and Downstream Applications},
year={2021},
url={https://openreview.net/forum?id=qw8AKxfYbI}
}

@InProceedings{lin2017fpn,
author = {Lin, Tsung-Yi and Dollar, Piotr and Girshick, Ross and He, Kaiming and Hariharan, Bharath and Belongie, Serge},
title = {Feature Pyramid Networks for Object Detection},
booktitle = {Proceedings of the IEEE Conference on Computer Vision and Pattern Recognition (CVPR)},
month = {July},
year = {2017},
pages = {2117-2125}
}

@InProceedings{sun2019hrnet,
author = {Sun, Ke and Xiao, Bin and Liu, Dong and Wang, Jingdong},
title = {Deep High-Resolution Representation Learning for Human Pose Estimation},
booktitle = {Proceedings of the IEEE/CVF Conference on Computer Vision and Pattern Recognition (CVPR)},
month = {June},
year = {2019},
pages = {5693-5703}
}
